\pdfoutput=1

\documentclass[11pt]{article}

\usepackage{ACL2023}

\usepackage{times}
\usepackage{latexsym}
\usepackage{bbm}

\usepackage[T1]{fontenc}

\usepackage[utf8]{inputenc}

\usepackage{microtype}

\usepackage{inconsolata}

\definecolor{my_green}{RGB}{51,102,0}
\definecolor{my_red}{RGB}{204, 0, 0}

\usepackage{graphicx}
\usepackage{array}
\usepackage{tabularx}
\usepackage[export]{adjustbox}
\usepackage{booktabs}
\usepackage{multirow}
\usepackage{caption}
\usepackage{xcolor}
\usepackage{colortbl}
\usepackage{amsmath}
\usepackage{mathtools}
\usepackage{pifont}
\usepackage{makecell}
\usepackage{hhline}
\usepackage{enumitem}
\usepackage{subcaption}
\usepackage{threeparttable}
\usepackage{algorithm}
\usepackage{listings}
\usepackage{algpseudocode}
\usepackage{lipsum}
\usepackage{hyperref}
\usepackage{tablefootnote}
\usepackage{marvosym}

\DeclareCaptionType{mycap}[Listing][List of Listings]

\lstdefinelanguage{JSON}{
    basicstyle=\ttfamily\footnotesize\color{black},
    numbers=none,
    numberstyle=\tiny\color{gray},
    stepnumber=1,
    numbersep=5pt,
    backgroundcolor=\color{white},
    showspaces=false,
    showstringspaces=false,
    showtabs=false,
    frame=single,
    tabsize=4,
    breaklines=true,
    breakatwhitespace=false,
    morestring=[b]",
    morecomment=[l]{//},
    morecomment=[s]{/*}{*/},
    keywordstyle=\color{blue},
    stringstyle=\color{black},
    commentstyle=\color{green},
    keywords={true,false,null}
}

\lstdefinestyle{algorithmstyle}{
    basicstyle=\ttfamily\footnotesize,
    escapeinside={(*@}{@*)}, 
    columns=fullflexible,
    frame=single,
    showstringspaces=false,
    keywordstyle=\color{blue},
    stringstyle=\color{red},
    commentstyle=\color{green},
}

\lstdefinestyle{jsonstyle}{
    basicstyle=\ttfamily\footnotesize,
    keywordstyle=\color{blue},
    stringstyle=\color{red},
    commentstyle=\color{green},
    numbers=left,
    numberstyle=\tiny\color{gray},
    stepnumber=1,
    numbersep=5pt,
    backgroundcolor=\color{white},
    showspaces=false,
    showstringspaces=false,
    showtabs=false,
    frame=single,
    tabsize=4,
    breaklines=true,
    breakatwhitespace=false,
    language=json
}

\definecolor{darkblue}{RGB}{94,110,186}

\newcommand{\cmark}{\textcolor{my_green}{\ding{51}}} 
\newcommand{\xmark}{\textcolor{my_red}{\ding{55}}} 
\newcommand{\darkblue}[1]{\textcolor{darkblue}{#1}}

\newcommand{\OurBenchmark}{AVUT}
\newcommand{\ManualBenchmark}{AV-Human}
\newcommand{\GeminiBenchmark}{AV-Gemini}

\title{Audio-centric Video Understanding Benchmark without Text Shortcut}

\author{
    Yudong~Yang$^{1,3*}$,
    Jimin~Zhuang$^{1,3*}$,
    Guangzhi~Sun$^{2,3}$, 
    Changli~Tang$^{1,3}$, 
    Yixuan~Li$^{1,3}$,
    \\ 
    \bf 
    Peihan~Li$^{3}$,
    Yifan~Jiang$^{3}$,
    Wei~Li$^{3}$, 
    Zejun~Ma$^{3}$, 
    Chao~Zhang$^{1}$\textsuperscript{\Letter}\vspace{2mm}
    \\ 
     \small
    \begin{tabular}{c} 
    $^1$Tsinghua University~~~~
    $^2$University of Cambridge~~~~
    $^3$ByteDance\vspace{2mm}\\
    \end{tabular}
    \\ 
    \small
    \begin{tabular}{c}
    {\{yang-yd21, zhuangjm21\}@mails.tsinghua.edu.cn},~~~~{cz277@tsinghua.edu.cn} \\
    \end{tabular}
}

\begin{document}
\maketitle

\renewcommand{\thefootnote}{}
\footnotetext{$^*$ Equal contribution; \ \ \Letter: Corresponding author.}
\renewcommand{\thefootnote}{\arabic{footnote}} 

\begin{abstract}
Audio often serves as an auxiliary modality in video understanding tasks of audio-visual large language models (LLMs), merely assisting in the comprehension of visual information. However, a thorough understanding of videos significantly depends on auditory information, as audio offers critical context, emotional cues, and semantic meaning that visual data alone often lacks. This paper proposes an audio-centric video understanding benchmark (\textbf{\OurBenchmark}) to evaluate the video comprehension capabilities of multimodal LLMs with a particular focus on auditory information. {\OurBenchmark} introduces a suite of carefully designed audio-centric tasks, holistically testing the understanding of both audio content and audio-visual interactions in videos. Moreover, this work points out the text shortcut problem that largely exists in other benchmarks where the correct answer can be found from question text alone without needing videos. \textbf{\OurBenchmark} addresses this problem by proposing a answer permutation-based filtering mechanism.
A thorough evaluation across a diverse range of open-source and proprietary multimodal LLMs is performed, followed by the analyses of deficiencies in audio-visual LLMs. Demos and data are available at \url{https://github.com/lark-png/AVUT}.

\end{abstract}

\section{Introduction}
\label{sec:intro}
\begin{table*}[h]
\centering
\resizebox{\textwidth}{!}{
    \begin{tabular}{lcccccccc}
    \toprule
    \multicolumn{1}{l}{\multirow{1}{*}{\textbf{Benchmarks}}} & \multicolumn{1}{c}{\multirow{1}{*}{\textbf{Num. of Videos}}} & \multicolumn{1}{c}{\multirow{1}{*}{\textbf{Avg.Duration(s)}}} & \multirow{1}{*}{\textbf{QA Pairs}} & \multicolumn{1}{c}{\multirow{1}{*}{\textbf{Tasks}}} & \multicolumn{1}{c}{\multirow{1}{*}{\textbf{Annotation}}} & \multicolumn{1}{c}{\multirow{1}{*}{\textbf{w. Audio}}} & \multicolumn{1}{c}{\multirow{1}{*}{\textbf{Ori. Videos}}} \\
    
    \midrule
    \multicolumn{8}{l}{\textit{\textbf{Video Benchmark}}} \\
    NExT-QA \citeyearpar{2021nextqa} & 1,000 & 39.5 & 8,564 & 3 & A & \xmark & \xmark  \\
    Video-Bench \citeyearpar{2023videobench} & 5,917 & 56.0 & 17,036 & 10 & SA & \xmark &  \cmark \\
    EgoSchema \citeyearpar{2023egoschema} & 5,063 & 180.0 & 5,063 & 1 & SA & \xmark & \xmark \\
    MVBench \citeyearpar{2024mvbench} & 3,641 & 16.0 & 3,641 & 20 & A & \xmark & \xmark \\
    MMBench-Video \citeyearpar{fang2024mmbench} & 609 & 165.4 & 1,998 & 26 & M & \xmark & \xmark    \\
    \midrule
    \multicolumn{8}{l}{\textit{\textbf{Audio-Visual Benchmark}}} \\
    LongVALE \citeyearpar{2024longvale} & 8,400 & 235.0 & - & 3 & SA & \cmark & \xmark   \\
    AVHBench \citeyearpar{2024avhbench} & 2,327 & - & 5,816 & 4 & SA & \cmark & \xmark    \\
    Video-MME \citeyearpar{2024videomme} & 900 & 1017.9 & 2,700 & 12 & M & \cmark & \cmark  \\
    \midrule
    \rowcolor[rgb]{ .851,  .906,  .992}
    \textbf{\OurBenchmark} & 2,662 & 67.8 & 11,609 & 8 & M+SA & \cmark & \cmark    \\
    \bottomrule
    \end{tabular}%
}
\caption{Comparison of {\OurBenchmark} with other representative video and audio-visual benchmarks. 
``Annotation'' is the annotation type (A: Automatic, SA: Semi-Automatic, M: Manual, M+SA: Mixed); ``w. Audio'' indicates the presence of audio; and ``Ori. Videos'' denotes whether the videos are newly collected (\cmark) or repurposed (\xmark).}
\label{tab:dataset_compare}
\vspace{-1em}
\end{table*}
Recent advances in Multimodal Large Language Models (MLLMs) \cite{2024gpt4o, 2024gemini1.5pro, Qwen2VL, 2024llavavideo, 2024longvila} have led to significant progress in video understanding. 
Evaluation of these models has been focusing predominantly on their visual abilities, with a large number of video benchmarks proposed in the past few years \cite{2021nextqa, 2023VideoChatGPT, 2023perceptiontest, 2024mvbench, 2023egoschema}. While several benchmarks \cite{avqa2022, UnAV2023, 2023vast, valor2023, 2024videomme, 2024longvale, 2024avhbench} incorporate audio in their evaluations, they are still largely visual-centric where the audio modality only acts as a secondary source of information. 

However, a comprehensive understanding of videos relies heavily on auditory information, as it provides essential context, emotional cues, and semantic meaning often absent from visual data alone. Existing benchmarks either overlooks the importance of audio in video understanding, or only focus on the speech content and ignoring other information. Consequently, there is an urgent need for dedicated efforts to evaluate model capabilities to understand and process speech and audio in videos. Moreover, most existing video understanding benchmarks suffer from the text shortcut problem, where the model can deduce the correct answer by simply focusing on the text alone. To fill the gaps, we propose \textbf{\OurBenchmark}, the text-shortcut-free benchmark specifically designed for audio-centric video understanding. The key differences of \textbf{\OurBenchmark} compared with examples of existing benchmarks are shown in Table \ref{tab:dataset_compare}.

\textbf{\OurBenchmark} introduces a suite of carefully designed audio-centric video understanding tasks, which cover not only content understanding of the video but also the interaction between audio and visual information, such as time synchronization or content matching. Besides, \textbf{\OurBenchmark} adopts a dual-dataset design which combines high-quality human annotation with scalable, semi-automatic annotation to obtain a holistic evaluation benchmark. Specifically, a novel Gemini-powered in-context learning data creation approach is proposed to generate high-quality question-answer pairs that are further validated by human annotators. \textbf{\OurBenchmark} incorporates \textbf{2,662} videos meticulously selected from YouTube, covering 18 audio-centric domains. This collection is split into two partitions: (1) \ManualBenchmark, comprising 698 videos with 1,734 expertly human-annotated question-answer pairs; and (2) \GeminiBenchmark, containing the remaining 1,964 videos with \textbf{9,875} question-answer pairs obtained from the proposed in-context learning data creation method. In addition, to address the text shortcut problem, a novel answer permutation-based filtering mechanism is proposed and applied in \textbf{\OurBenchmark}, ensuring the necessity of multimodal inputs.
The main contributions are as follows:

\begin{itemize}[leftmargin=*]
    \setlength\itemsep{-0.2em}
    \item We introduce \textbf{\OurBenchmark}, the \textbf{A}udio-centric \textbf{V}ideo \textbf{U}nderstanding benchmark without \textbf{T}ext shortcut, featuring diverse tasks encompassing both audio content understanding and audio-visual alignment.
    \item \textbf{\OurBenchmark} first systematically investigate the text shortcut problem in existing audio and video benchmarks, and proposes a novel filtering mechanism to minimize the text shortcut and ensure sufficient necessity of the multimodal inputs.
    \item We evaluate a diverse range of audio-visual, visual-only, and audio-only models on \textbf{{\OurBenchmark}} to demonstrate their capabilities on audio-centric video understanding.
\end{itemize}

\section{Related Work}
\paragraph{MLLMs.} Early research in multimodal alignment explored the creation of shared representations across images, audio, and text \cite{aytar2017see}, demonstrating the potential of transferable classifiers. Further research on shared representations across modalities leads to encoders that can align with LLMs like Whisper \cite{whisper} or CLIP \cite{clip}, making MLLMs technically possible. Recent progress in audio-visual learning has spurred the development of large language models (LLMs) capable of processing both audio and visual information. Early video understanding works like VideoChat \cite{2023videochat} uses external subtitles for speech. Subsequent models have explored approaches to fuse audio and visual modalities. PandaGPT \cite{PandaGPT} integrates ImageBind \cite{2023imagebind} and Vicuna \cite{vicuna2023}, leveraging image-text pairings for enhanced cross-modal understanding, while VAST \cite{2023vast}, trained on the large-scale VAST-27M \cite{2023vast} dataset, further advances multimodal capabilities. Models like CAT \cite{2025cat} employ specialized architectures, such as clue aggregators, to effectively process audio-visual events. Video-LLaMA \cite{2023videollama}, VideoLLaMA2 \cite{2024videollama2} and Baichuan Omni \cite{baichuanomni} have been fine-tuned specifically on audio data to improve multimodal comprehension. Architectures like the multi-resolution causal Q-former in video-SALMONN \cite{videosalmonn} aim to bridge pre-trained audio-visual encoders with LLMs efficiently. Furthermore, models like NExTGPT \cite{2024nextgpt}, OneLLM \cite{2023onellm}, and VITA \cite{2024vita} use ``omni-encoders'' to uniformly process various data modalities, including audio and video.

\paragraph{Video Benchmark.} Numerous efforts have focused on developing video benchmarks \cite{2021nextqa, 2021value, 2024mvbench, 2023videobench, 2023vitatecs, 2024tempcompass, 2023egoschema, 2024videomme, fang2024mmbench}, predominantly targeting the visual aspects of video content. However, the crucial role of audio in video understanding has received comparatively less attention. Several early datasets explore the use of audio in video understanding. AVSD \cite{AVSD2019} focuses on scene-aware dialogue, Pano-AVQA \cite{pano-AVQA2021} on panoramic videos, MUSIC-AVQA \cite{musicAVQA2022} specifically on musical contexts, and VGGSound \cite{2020vggsound} on short videos of audio events. Thus, these datasets target specific scenarios. While AVQA \cite{avqa2022} explores the synergy between audio and visual information but is limited by shorter video lengths and simplistic task design.
More recent datasets like UnAV-100 \cite{UnAV2023}, VALOR \cite{valor2023}, and VAST-27M \cite{2023vast} offer large-scale data but prioritize specific tasks, such as audio event detection, retrieval, and captioning. Similarly, LongVALE \cite{2024longvale} focuses on audio-visual-language events, and AVHBench \cite{2024avhbench} on cross-modal hallucination. These benchmarks, often lacking human annotation, are less suited for evaluating the nuanced and complex aspects of audio-centric understanding, particularly alignment capabilities,  targeted by \OurBenchmark.

\section{Our Benchmark}
\subsection{Data Curation Principles}
\textbf{Focusing on Audio-Centric Understanding in Video.} Unlike existing video benchmarks that predominantly target visual understanding, {\OurBenchmark{}} focuses specifically on the crucial role of audio in video comprehension. To this end, the tasks in {\OurBenchmark{}} are designed to focus on the following two aspects: (1) \textbf{Audio Content Understanding}, requiring models to perceive and understand audio information; (2) \textbf{Audio Visual Alignment}, requiring the models to align audio with the corresponding visual information. 

\textbf{Covering Diverse Audio-Centric Video Domains and Comprehensive Annotation.} To thoroughly evaluate audio-centric understanding capabilities in MLLMs, {\OurBenchmark{}} incorporates a diverse selection of 18 video domains where audio plays a central role. For robust evaluation, {\OurBenchmark{}} offers both \textbf{\ManualBenchmark{}}, a meticulously human-annotated dataset, and \textbf{\GeminiBenchmark{}}, a significantly larger dataset generated via a novel Gemini-augmented approach. This combination of diverse content and rich annotation enables {\OurBenchmark{}} to effectively support the research in audio-centric video understanding.

\subsection{Audio-Centric Task Definition}
\begin{table*}[tp]
    \centering
    \setlength\tabcolsep{4pt}
    \resizebox{1.0\textwidth}{!}{
        \begin{tabular}{c|c|c|c}
        \Xhline{1.0pt}
        \textbf{Theme} & \textbf{Task Type} & \textbf{Format} & \textbf{Example} \\
        \Xhline{1.0pt}
        \multirow{6}{*}{\textbf{Audio Content Understanding}} & \multirow{2}{*}{Audio Information Extraction} & \multirow{2}{*}{Multiple Choice} & \cellcolor{gray!5}{\textit{\darkblue{The person in the audio said when they would have a trip from Central Africa to the farthest edge of Norway?}}} \\
        ~ & ~ & ~ & \cellcolor{gray!5}{(A) The spring of 2025. (B) The summer of 2025. (C) The spring of 2026. (D) The summer of 2026.} \\
        \hhline{~|-|-|-}
        ~ & \multirow{2}{*}{Audio Content Counting} & \multirow{2}{*}{Multiple Choice} & \cellcolor{gray!5}{\textit{\darkblue{How many times does the rumbling sound appear in the video?}}} \\
        ~ & ~ & ~ & \cellcolor{gray!5}{(A) 2. (B) 3. (C) 4. (D) 5.} \\
        \hhline{~|-|-|-}
        ~ & \multirow{2}{*}{Audio Event Localization} & \multirow{2}{*}{Multiple Choice} & \cellcolor{gray!5}{\textit{\darkblue{In what seconds does the person begin to sing In the video?}}}  \\
        ~ & ~ & ~ & \cellcolor{gray!5}{(A) 8 to 9. (B) 10 to 11. (C) 13 to 14. (D) 15 to 16.} \\
        \Xhline{1.0pt}
        \multirow{14}{*}{\textbf{Audio Visual Alignment}} & \multirow{3}{*}{Audio Visual Character Matching} & \multirow{3}{*}{Multiple Choice} & \cellcolor{gray!5}{\textit{\darkblue{Who says "there's no way of knowing who did it" in the video?}}} \\
        ~ & ~ & ~ & \cellcolor{gray!5}{(A) The woman on the bottom right of the screen. (B) The man on the top right of the screen.} \\
        ~ & ~ & ~ & \cellcolor{gray!5}{(C) The man on the top left of the screen. (D) The woman on the bottom left of the screen.} \\
        \hhline{~|-|-|-}
        ~ & \multirow{2}{*}{Audio Visual Object Matching} & \multirow{2}{*}{Multiple Choice} & \cellcolor{gray!5}{\textit{\darkblue{What happens when the singer in the video finishes singing "talk to me"?}}} \\
        ~ & ~ & ~ & \cellcolor{gray!5}{(A) A picture appears. (B) A man appears. (C) A woman appears. (D) Nothing.} \\
        \hhline{~|-|-|-}
        ~ & \multirow{3}{*}{Audio Visual Text Matching} & \multirow{3}{*}{Multiple Choice} & \cellcolor{gray!5}{\textit{\darkblue{When the number 48KG appears on the screen, what is being explained at this time?}}} \\
        ~ & ~ & ~ & \cellcolor{gray!5}{(A) J2 Weight requirements for competitors. (B) Classification of judo.} \\
         ~ & ~ & ~ & \cellcolor{gray!5}{(C) Judo techniques. (D) The duration of a judo competition.} \\
         \hhline{~|-|-|-}
        ~ & \multirow{2}{*}{Audio Visual Segment Matching} & \multirow{2}{*}{Open Ended} & \cellcolor{gray!5}{\textit{\darkblue{The model is prompted to sort the audio pieces of a video whose audio was clipped into 4 pieces and shuffled.}}} \\
        ~ & ~ & ~ & \cellcolor{gray!5}{Success rate of sorting the entire audio and any two clips is calculated.} \\
        \hhline{~|-|-|-}
        ~ & \multirow{2}{*}{Audio Visual Speaker Dirization} & \multirow{2}{*}{Open Ended} & \cellcolor{gray!5}{\textit{\darkblue{The model is instructed to transcribe from only the speaker with a certain visual characteristic.}}} \\
        ~ & ~ & ~ & \cellcolor{gray!5}{WER calculation is then applied to the model's transcription.} \\
        \Xhline{1.0pt}
        \end{tabular}
        }
    \caption{Task examples of {\OurBenchmark}. 
    ``Audio Visual Segment Matching'' is to sort four randomly shuffled audio segments in a video into their original order. ``Audio Visual Speaker Diarization'' is to transcribe the speech of a visually specified speaker with the Word Error Rate (WER) as the metric.}
    \label{tab:task_example}
    \vspace{-0.3cm}
\end{table*}

The tasks are designed to probe specific aspects of audio-centric video understanding, presenting unique challenges for MLLMs. Examples of each task are shown in Table \ref{tab:task_example}.

\subsubsection{Audio Content Understanding}
The ability to perceive and process audio is fundamental to audio-centric video understanding. Inspired by classic audio processing challenges, we designed tasks to probe a model’s capability to extract meaningful information, quantify audio events, and locate them in a video. Therefore, we define three tasks as follows:
\begin{itemize}[leftmargin=*]
\setlength\itemsep{-0.3em}
    \item \textbf{Audio Information Extraction (AIE)} is to extract specific information from spoken language, such as sentiment, key facts, or main points.
    \item \textbf{Audio Content Counting (ACC)} is to test the accuracy of counting the number of occurrences for a specific audio event. 
    \item \textbf{Audio Event Location (AEL)} is to pinpoint the precise start and end time of a specific audio event in the video.
\end{itemize}

\subsubsection{Audio Visual Alignment}
Existing benchmarks often lack a dedicated focus on this cross-modal alignment, which is fundamental to a comprehensive understanding of video content. To bridge this gap, we define five matching-based tasks as follows.

\begin{itemize}[leftmargin=*]
\setlength\itemsep{-0.2em}
    \item \textbf{Audio-Visual Character Matching (AVCM)} tests whether the model links spoken content to the corresponding character's visual appearance.
    \item \textbf{Audio-Visual Object Matching (AVOM)} is to draw the connection between utterances and depicted objects or background scenes. 
    \item \textbf{Audio-Visual Text Matching (AVTM)} requires the model to associate spoken information with on-screen text, e.g. subtitles.
    \item \textbf{Audio-Visual Segment Matching (AVSM)} tests whether the model aligns audio clips with their corresponding video segments. 
    \item \textbf{Audio-Visual Speaker Diarization (AVDiar)} requires the model to determine ``who spoke when'' by integrating audio and visual cues.
\end{itemize}

\subsection{Dataset Construction}
{\OurBenchmark} contains \textbf{2,662} videos with \textbf{11,609} questions in total with each question either created or validated by human annotators. This section describes the data construction in detail.

\subsubsection{Video Collection}

\begin{figure*}[t]
    \centering
    \includegraphics[width=0.95\textwidth]{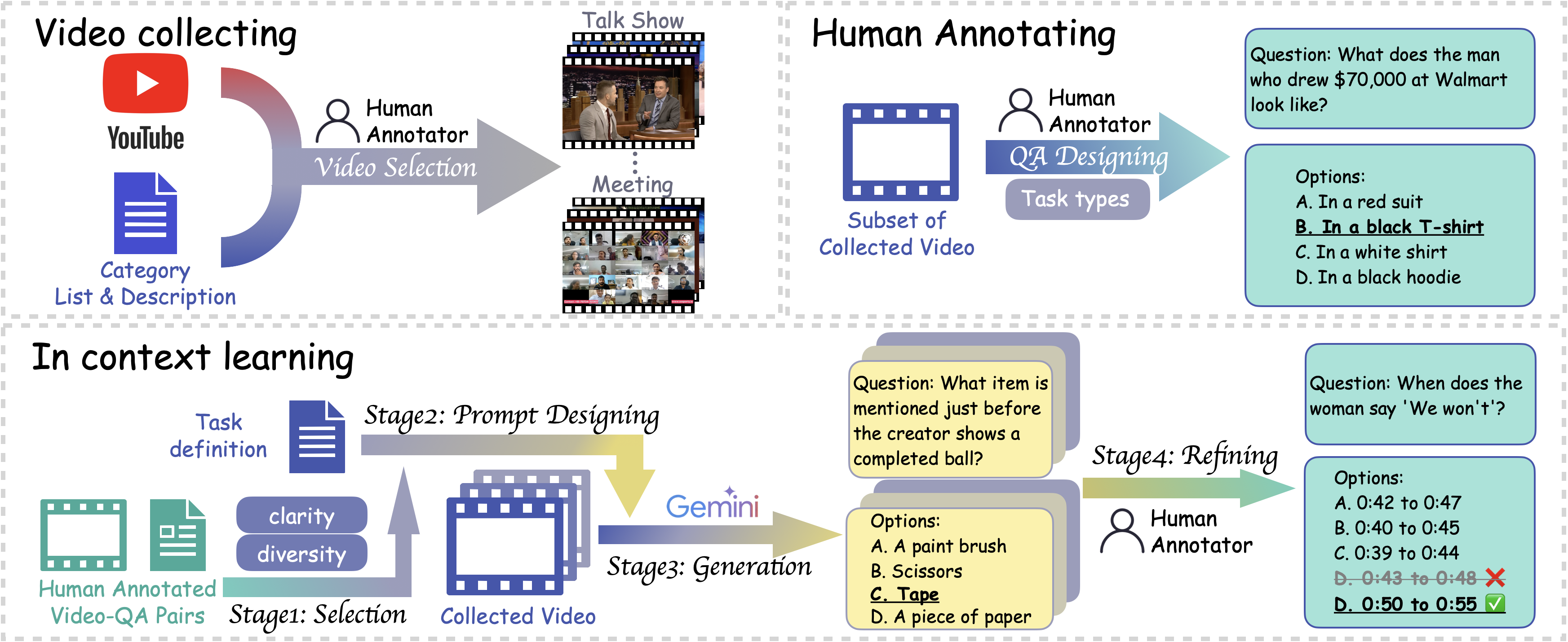}
    \caption{The Pipeline of {\OurBenchmark}. The pipeline consists of three main parts: video collection, human annotation and in-context learning creation. Annotators select videos from YouTube and these videos are then used for human annotation, where question-answer pairs are designed based on various task types. The remaining videos are used for in-context learning with Gemini 1.5 Pro. The in-context learning process involves four stages: (1) Demonstration Selection; (2) Prompt Designing; (3) Generation; (4) Refining.}
    \vspace{-0.2cm}
    \label{fig:pipeline}
\end{figure*}

We define 18 audio-centric video domains (see Appendix \ref{appendix example videos and tasks}) to cover a diverse range of scenarios in which audio plays a crucial role in comprehension. Our annotators searched YouTube for videos within these audio-centric domains to manually collect a total of 2,662 raw videos, as depicted in Figure \ref{fig:pipeline}. The collected videos were required to be in English, rich in audio content, and recently published (mostly after 2023) to minimize overlap with existing training sets. Videos were intentionally limited to two minutes as driven by the following considerations: First, the nature of our audio-centric tasks, which focus on analyzing specific audio events or aligning audio cues with corresponding visual information within localized segments, lends itself well to evaluation within this shorter time frame. Second, most popular open-source MLLMs struggle to process longer videos, and hence short videos allow us to perform a wider range of comparisons. Finally, extending the length of the video would lead to an exponential increase in the cost and potential inaccuracies of manual annotation, without significantly enhancing the discriminatory power of our benchmark. These videos further went through a validation process, where independent annotators confirmed the designated domains and the presence of sufficient and relevant audio information. The detailed video collection process is provided in Appendix \ref{appendix: video collection}.

\subsubsection{Human Annotation Partition}
\label{section:annotation}

The video collection and annotation for {{\ManualBenchmark}} was conducted by a team of trained annotators with expertise in data labeling and rigorous quality control procedures (see Appendix \ref{appendix:annotation} for details). 

\textbf{Annotation Procedure.} We selected 698 videos for full annotation where annotators were given detailed task definitions and annotation examples.
Then, for each video, annotators selected the three most suitable task types and created corresponding multiple-choice questions and answer options, resulting in a total of \textbf{1,734} question-answer pairs. For AVDiar, annotators selected 260 videos from the full annotation set that contain a higher number of speakers (3.59 speakers per video on average). Detailed annotations for these videos included speaker appearance, start and end times of utterances, and speech transcriptions. The Audio-Visual Matching task did not require manual annotation, as audio-video segment pairings were generated automatically using a script.

\textbf{Quality Assurance.} For each task in \OurBenchmark, we selected and trained annotators with relevant expertise. A rigorous two-stage quality control process, involving specialist review and cross-validation, ensured high-quality annotations. During annotation, questions were carefully designed to require genuine multimodal understanding, avoiding common-sense reasoning and text-based answers.

\subsubsection{Gemini-Augmented Partition}

Human annotations are both costly and often have a bias towards visual-grounded questions (see Appendix \ref{appendix: annotation bias}). 
To mitigate these limitations and to expand to a larger scale, we employed a semi-automatic annotation strategy using Gemini 1.5 Pro and created the {\GeminiBenchmark} subset. Specifically, we propose an \textbf{in-context learning} approach which includes 4 stages as shown in Figure \ref{fig:pipeline}:

\textbf{Stage 1: Demonstration Selection.} We carefully selected four representative examples from the human-annotated data for each of the six multiple-choice question task types. These examples, chosen for their clarity and diversity, served as high-quality demonstrations for Gemini.

\textbf{Stage 2: Prompt Construction. }We constructed prompts for Gemini by combining the task definition with the selected demonstrations. Each prompt provided clear instructions and illustrative examples to guide Gemini's generation process.

\textbf{Stage 3: Automated Annotation Generation. }We provided Gemini with the remaining 1,964 raw videos from our collected set of 2,662. For each video and each task type, Gemini generates a question, four options and the correct option, resulting in a total of \textbf{9,875} generated QA pairs.

\textbf{Stage 4: Human Review and Refinement.} All generated QA pairs underwent meticulous human review, focusing on accuracy, relevance, adherence to task requirements, and answer correctness. Annotators manually corrected any discrepancies. (See Appendix \ref{appendix: Gemini refinement} for correction examples.)

\section{Text Shortcut Problem}
Text shortcut occurs when models give correct answers by only looking at the textual content of the question without requiring the corresponding video, which results in unreliable evaluation of the video understanding capabilities. This section analyzes the causes and consequences of text shortcut and details the methodology employed in {\OurBenchmark} to mitigate this issue.

\textbf{Text shortcut arises from several factors.} Bias in question formulation can inadvertently provide textual cues or biases that hint at the correct answer, even without viewing the video. For instance, a question about a "fast-moving animal" might bias the model towards choosing "cheetah" even if the video depicts a different animal. Furthermore, models may have encountered similar question-answer pairings during their training, effectively memorizing the associations rather than learning to reason about video content. Finally, the powerful text processing capabilities of advanced language models like GPT-4o allow them to exploit subtle textual cues that humans annotators might overlook.

\textbf{Text shortcut severely compromises evaluation validity.} Models may appear to perform better than they genuinely do, masking their true video understanding capabilities and leading to inflated performance metrics. This can, in turn, misdirect research efforts towards optimizing for textual cues rather than genuine video comprehension, hindering true progress in multimodal learning.

\textbf{{\OurBenchmark} adopts a filtering mechanism based on answer permutation.} For each multiple-choice question, we created four cyclic permutations of the answer options (ABCD, BCDA, CDAB, DABC). The permutations help exclude the influence of potential positional bias in the LLM. We then present these variations of the same question to GPT-4o and check the correctness under each permutation. Without video inputs, if the model can still find the correct answer under all four permutations, the question is likely to have a text shortcut. These questions, deemed solvable through textual cues alone, were subsequently removed from {\OurBenchmark}. This filtering mechanism ensures that the video content is essential to finding the correct answer in the remaining questions.

\section{Experiment}
\begin{table*}[t]
\centering
\begin{adjustbox}{max width=\linewidth}
\small
\begin{tabular}{l!{\vrule width \lightrulewidth}cccccc} 
\toprule
\multicolumn{1}{c|}{\multirow{1}{*}{\textbf{Models}}}  & \multicolumn{1}{c}{\textbf{AIE (\%)}} & \multicolumn{1}{c}{\textbf{ACC (\%)}} & \multicolumn{1}{c}{\textbf{AEL (\%)}} & \multicolumn{1}{c}{\textbf{AVCM (\%)}} & \multicolumn{1}{c}{\textbf{AVOM (\%)}} & \multicolumn{1}{c}{\textbf{AVTM (\%)}} \\ 

\midrule
\multicolumn{7}{c}{\textit{Audio Visual MLLMs}} \\ 
\midrule

Gemini 1.5 Pro \citeyearpar{2024gemini1.5pro} & \textbf{90.02} / \textbf{94.70} & \textbf{50.85} / \textbf{39.99} & \textbf{83.43} / \textbf{73.32} & \textbf{74.05} / \textbf{83.13} & \textbf{74.79} / \textbf{76.62} & \textbf{81.22} / \textbf{83.68} \\
VideoLLaMA2 (7B) \citeyearpar{2024videollama2} & 42.30 / 35.65 & 32.77 / 32.88 & 32.02 / 25.75 & 46.80 / 57.41 & 49.60 / 46.11 & 48.92 / 40.90  \\
video-SALMONN (13B) \citeyearpar{videosalmonn} & 52.45 / 32.86 & 33.33 / 30.17 & 36.05 / 32.48 & 34.47 / 41.42 & 30.25 / 31.42 & 39.95 / 36.21 \\ 
PandaGPT (13B) \citeyearpar{PandaGPT} & 24.94 / 30.82 & 25.00 / 29.13 & 26.16 / 30.61 & 24.31 / 27.54 & 26.28 / 28.44 & 25.77 / 26.36   \\

\midrule
\multicolumn{7}{c}{\textit{Visual MLLMs}} \\ 
\midrule

GPT-4o \citeyearpar{2024gpt4o} & \textbf{60.37} / \textbf{61.43} & 35.00 / 31.14 & 33.14 / 34.57 & 51.81 / 55.71 & 59.34 / 55.43 & \textbf{70.69} / \textbf{61.71}     \\
Qwen2-VL (7B) \citeyearpar{Qwen2VL} & 59.21 / 53.48 & \textbf{41.67} / \textbf{33.18} & \textbf{34.88} / \textbf{39.44} & \textbf{62.90} / \textbf{65.08} & 55.85 / 57.01 & 69.74 / 58.37 \\
LLaVA-Video (7B) \citeyearpar{2024llavavideo} & 54.19 / 55.77 & 39.05 / 30.62 & 24.83 / 38.17 & 62.83 / 62.53 & \textbf{61.87} / \textbf{58.21} & 62.94 / 60.04   \\
InternVL2 (76B) \citeyearpar{internvl2} & 53.15 / 45.49 & 25.83 / 26.83 & 30.23 / 26.89 & 58.42 / 57.79 & 53.18 / 50.13 & 61.70 / 53.25  \\
InternVL2 (8B) \citeyearpar{internvl2} & 44.52 / 37.70 & 30.00 / 25.24 & 29.07 / 26.58 & 49.25 / 53.54 & 47.43 / 43.38 & 53.19 / 43.69   \\
VILA-1.5 (8B) \citeyearpar{2024vila} & 41.72 / 38.28 & 28.33 / 26.78 & 28.49 / 36.61 & 53.30 / 62.38 & 46.41 / 47.77 & 46.34 / 46.64    \\
VideoLLaVA (7B) \citeyearpar{2023videollava} & 34.73 / 27.97 & 15.00 / 19.10 & 22.67 / 17.30 & 36.67 / 42.36 & 34.09 / 35.86 & 35.93 / 29.69 \\

\midrule
\multicolumn{7}{c}{\textit{Audio MLLM}} \\ 
\midrule

SALMONN (13B) \citeyearpar{2024salmonn} & 45.92 / 45.18 & 33.33 / 31.68  & 29.65 / 31.56 & 32.62 / 32.70 & 36.76 / 33.11 & 34.52 / 38.39  \\

\bottomrule

\end{tabular}
\end{adjustbox}
\caption{Task Results (Multiple-Choice Questions). This table presents the performance of each model on the multiple-choice question tasks. AIE: Audio Information Extraction, ACC: Audio Content Counting, AEL: Audio Event Localization, AVCM: Audio-Visual Character Matching, AVOM: Audio-Visual Object Matching, AVTM: Audio-Visual Text Matching. Results are presented as accuracy scores (\%). Each cell shows the performance on AV-Human (left) and AV-Gemini (right) datasets, separated by a forward slash (/).}
\label{tab:Task result}
\end{table*}

\begin{table}[t]
\centering
\begin{adjustbox}{max width=\linewidth}

\begin{tabular}{l!{\vrule width \lightrulewidth}c!{\vrule width \lightrulewidth}c!{\vrule width \lightrulewidth}c} 
\toprule
\multicolumn{1}{c|}{\multirow{1}{*}{\textbf{Models}}} & \multicolumn{1}{c|}{\textbf{\ManualBenchmark}} & \multicolumn{1}{c|}{\textbf{\GeminiBenchmark}} & \multicolumn{1}{c}{\textbf{Overall}}  \\ 

\midrule
\multicolumn{4}{c}{\textit{Audio Visual MLLMs}} \\ 
\midrule

Gemini 1.5 Pro \citeyearpar{2024gemini1.5pro} & 78.34 & 75.21 & 75.67 \\
VideoLLaMA2 (7B) \citeyearpar{2024videollama2} & 44.90 & 39.75 & 40.56 \\ 
video-SALMONN (13B) \citeyearpar{videosalmonn} & 38.33 & 34.08 & 34.74 \\
PandaGPT (13B) \citeyearpar{PandaGPT} & 25.38 & 28.83 & 28.31 \\

\midrule
\multicolumn{4}{c}{\textit{Visual MLLMs}} \\ 
\midrule

GPT-4o \citeyearpar{2024gpt4o}  & 56.62 & 50.00\tablefootnote{We evaluated GPT-4 on a subset of 2100 questions randomly sampled from the full question set with balanced task types, sufficiently representing the original task distribution.} & 53.31 \\
Qwen2-VL (7B) \citeyearpar{Qwen2VL} & 58.38 & 51.05 & 52.26 \\
LLaVA-Video (7B) \citeyearpar{2024llavavideo} & 56.52 & 50.62 & 51.48 \\
InternVL2 (76B) \citeyearpar{internvl2} & 52.62 & 43.29 & 44.72  \\
InternVL2 (8B) \citeyearpar{internvl2} & 45.90 & 38.31 & 39.47 \\
VILA-1.5 (8B) \citeyearpar{2024vila} & 44.48 & 43.05 & 43.27  \\
VideoLLaVA (7B) \citeyearpar{2023videollava} & 33.14 & 28.70 & 29.37  \\

\midrule
\multicolumn{4}{c}{\textit{Audio MLLM}} \\ 
\midrule

SALMONN (13B) \citeyearpar{2024salmonn} & 36.48 & 35.43 & 35.59   \\

\bottomrule

\end{tabular}
\end{adjustbox}
\caption{Overall Performance on {\OurBenchmark}. This table presents the overall performance of each model on the AV-Human, AV-Gemini, and Overall (combined AV-Human and AV-Gemini) datasets. Performance is measured by accuracy (\%) on MCQs.}
\label{tab:overall result}
\end{table}

\subsection{Experimental Setup}
We evaluated a range of prominent and representative models on {\OurBenchmark}, encompassing audio-visual, visual-only, and audio-only architectures, to comprehensively assess performance across different modalities. \textbf{Audio-Visual MLLMs:} We evaluated Gemini 1.5 Pro \cite{2024gemini1.5pro}, Video-SALMONN(13B) \cite{videosalmonn}, VideoLLaMA2(7B) \cite{2024videollama2}, and PandaGPT(13B) \cite{PandaGPT}. These models received both video and audio inputs. \textbf{Visual MLLMs:} We evaluated GPT-4o \cite{2024gpt4o}, Qwen-2-VL(7B) \cite{Qwen2VL}, LLaVA-video(7B) \cite{2024llavavideo}, InternVL2(76B and 8B) \cite{internvl2}, VILA-1.5(8B) \cite{2024longvila}, and VideoLLaVA(7B) \cite{2023videollava}. These models received video input with the audio track removed. \textbf{Audio MLLM:} We evaluated SALMONN(13B) \cite{2024salmonn} using only the audio track extracted from the videos.

For the closed-source models, GPT-4 and Gemini 1.5 Pro, we used the API versions ``gpt-4o-2024-05-13'' and ``gemini-1.5-pro-preview'', respectively, with a sampling rate of 1 frame per second (fps) and a temperature of 1.0. For open-source models, we adhered to their official configurations, including default parameter settings, decoding strategies, temperature settings, and frame sampling rates\footnote{Number of frames or frame rates are: Video-SALMONN, 2 fps; VideoLLaMA2, 16 frames total; PandaGPT, 10 frames total; Qwen2-VL, 1 fps; LLaVA-Video, 32 frames total; InternVL2, 32 frames total; VILA-1.5, 6 frames total; VideoLLaVA, 8 frames total; and SALMONN, 16kHz audio.}. During testing, audio-visual models received videos with audio, visual-only models received videos with the audio removed, and audio-only models received only the audio track extracted from the videos. All experiments were conducted on A100 GPUs. Experiments with closed-source models, utilizing their respective APIs, took an average of 6 hours to complete. Inference with open-source models required an average of 4 hours.

\subsection{Results}
Table \ref{tab:overall result} presents the overall performance of all evaluated models on the {\ManualBenchmark}, {\GeminiBenchmark}, and combined Overall ({\ManualBenchmark} + {\GeminiBenchmark}) datasets. Performance is measured by accuracy on the multiple-choice question tasks. To provide a more detailed analysis, Table \ref{tab:Task result} presents the per-task accuracy of all models on the {\ManualBenchmark}, {\GeminiBenchmark}. This result leads to further discussion on how the annotation method can influence the annotation result, and how different abilities of a model can be revealed through different task types.

\paragraph{AV-Human compared to AV-Gemini.} Most models show a performance drop when switching from AV-Human to AV-Gemini, particularly in audio content understanding tasks. However, they perform better on AV-Gemini in audiovisual alignment tasks due to differences in annotation methods.
Gemini, leveraging in-context learning, captures finer details for QA pair creation, while human annotators focus on a video's main theme, often missing subtleties and favoring visually dominant content. This makes human annotations less challenging for visual-only models.
In contrast, human annotations excel in audiovisual matching tasks due to access to full-frame-rate video and synchronized audio, enabling the creation of fine-grained therefore more challenging questions, which are more difficult for models to handle.
\paragraph{The difficulty of different tasks varies.} Most open-source models perform well in audio-visual character-matching, as human figures are visually prominent. However, their performance drops in audio-essential tasks like content counting and event localization, highlighting reliance on audio, showcased in case studies in Appendix \ref{appendix:case study}.
Nevertheless, visual models perform well on audio information extraction tasks due to the high correlation between audio and visual content. Such correlation enables models to leverage visual cues to answer questions, even in tasks designed to depend primarily on audio. This explains why visual-only models also perform well on the matching tasks, since the audio content can be inferred from visual to help solve the question. Besides, the audio-only model, SALMONN, has a more consistent performance among tasks, indicating a stable reliance on audio information across different tasks.

\paragraph{{\OurBenchmark} is designed to be a multi-modal benchmark where audio plays a crucial role.} To support this, we choose the most competitive audio-visual model, Gemini 1.5 Pro, to perform ablation studies as shown in Table \ref{tab:Modality ablation}. We compared the model's performance across several input configurations: full audio-visual, video-only, audio-only, and a detailed transcription setting. The "detailed transcription" includes speaker identification, corresponding ASR transcripts, and start/end timestamps for each utterance. This result demonstrated that our benchmark relies heavily on audio modality, as its performance drops significantly when missing any of the two modalities. While providing detailed transcriptions with speaker diarization and timestamps improves performance compared to video-only, it still falls short of the performance achieved with full audio. This suggests that beyond the transcribed words themselves, the rich information encoded in audio, such as intonation, prosody, and overlapping speech, is crucial for understanding the video content and is not fully captured even by detailed textual transcriptions. Transcription annotation examples are shown in Appendix \ref{appendix: Gemini transcription}.

\paragraph{Most audio-visual models underperform visual-only models even on audio-visual alignment tasks.} For weaker models, having audio-visual inputs may instead introduce more hallucinations, causing them to confuse visual-only and audio-only distractors, whereas video-only models only deal with visual distractors.
\begin{table}[t]
\centering
\begin{adjustbox}{max width=\linewidth}
\small
\begin{tabular}{l!{\vrule width \lightrulewidth}c!{\vrule width \lightrulewidth}c!{\vrule width \lightrulewidth}c!{\vrule width \lightrulewidth}c} 
\toprule
\multicolumn{1}{c|}{\multirow{1}{*}{\textbf{Modalities}}} & \multicolumn{1}{c|}{\textbf{Audio-Visual}}  & \multicolumn{1}{c|}{\textbf{Transcription}} & \multicolumn{1}{c|}{\textbf{Visual-Only}} & \multicolumn{1}{c}{\textbf{Audio-Only}} \\ 
\midrule
\textbf{Accuracy (\%)} & 78.34 & 69.23 & 62.78 & 54.10 \\

\bottomrule
\end{tabular}
\end{adjustbox}
\caption{Gemini 1.5 Pro's performance when provided different modalities.}
\label{tab:Modality ablation}
\vspace{-0.3cm}
\end{table}

\begin{table}[t]
\centering
\begin{adjustbox}{max width=\linewidth}

\begin{tabular}{l!{\vrule width \lightrulewidth}cc!{\vrule width \lightrulewidth}c} 
\toprule
\multicolumn{1}{c|}{\multirow{2}{*}{\textbf{Models}}} & \multicolumn{2}{c|}{\textbf{AVSM}} & \multicolumn{1}{c}{\textbf{AVDiar}} \\
\cmidrule(lr){2-3}\cmidrule(lr){4-4}
& Pair (\%) $\uparrow$  & Full (\%) $\uparrow$  & DWER (\%) $\downarrow$  \\

\midrule

Gemini 1.5 Pro \citeyearpar{2024gemini1.5pro} & 72.84 & 26.29 & 66.31 \\
VideoLLaMA2 (7B) \citeyearpar{2024videollama2} & 38.05 & 0.00 & 116.92 \\ 
PandaGPT (13B) \citeyearpar{PandaGPT} & 39.87 & 1.22 & 155.94 \\

\bottomrule

\end{tabular}
\end{adjustbox}
\caption{Open-Ended Task Results on Audio-Visual Segment Matching (AVSM) and Audio-Visual Speaker Diarization (AVDir). ``Pair ↑'' denotes the percentage of correctly sequenced segment pairs (not necessarily adjacent), ``Full ↑'' denotes the percentage of fully matched videos, ``DWER ↓'' represents Word Error Rate (WER) between ground truth and diarized transcriptions. }
\label{tab:AVSM and AVDir}
\end{table}

\paragraph{Open-ended QA tasks need stronger models.} As shown in Table \ref{tab:AVSM and AVDir}, although segmentation and diarisation are popular speech processing tasks, only Gemini 1.5 Pro achieved sensible results while other models failed to perform those tasks. This is partly due to the poor instruction-following abilities of audio-LLMs, which highlights an aspect that audio-visual LLMs need improvements on.

\paragraph{MCQs are capable enough for revealing models' ability.} We turned MCQs in {\ManualBenchmark} into open-ended questions, where the original correct option is used as the reference answer. The generated answer was ranked using GPT-4o and Rouge-L. Performance comparison between open-ended questions and MCQs in Table \ref{tab:Openended Ablation} showed that models performed similarly with different question formats. Despite differences from real-world scenarios, the MCQs still effectively reflect models' performance on audio-centric video understanding in real life.


\begin{table}[t]
\centering
\begin{adjustbox}{max width=\linewidth}

\begin{tabular}{l!{\vrule width \lightrulewidth}c!{\vrule width \lightrulewidth}c!{\vrule width \lightrulewidth}c} 
\toprule
\multicolumn{1}{c|}{\multirow{1}{*}{\textbf{Models}}} & \multicolumn{1}{c|}{\textbf{GPT rank}} & \multicolumn{1}{c|}{\textbf{Rouge-L}} & \multicolumn{1}{c}{\textbf{MCQ Acc}}  \\ 

\midrule
\multicolumn{4}{c}{\textit{Audio Visual MLLMs}} \\ 
\midrule

Gemini 1.5 Pro \citeyearpar{2024gemini1.5pro} & 3.331 & 0.108 & 78.34 \\
VideoLLaMA2 (7B) \citeyearpar{2024videollama2} & 1.839 & 0.062 & 44.90 \\ 
video-SALMONN (13B) \citeyearpar{videosalmonn} & 2.032 & 0.080 & 38.33 \\
PandaGPT (13B) \citeyearpar{PandaGPT} & 1.384 & 0.035 & 25.38 \\

\midrule
\multicolumn{4}{c}{\textit{Visual MLLMs}} \\ 
\midrule

Qwen2-VL (7B) \citeyearpar{Qwen2VL} & 2.233 & 0.074 & 58.38 \\
LLaVA-Video (7B) \citeyearpar{2024llavavideo} & 2.217 & 0.099 & 56.52 \\
InternVL2 (76B) \citeyearpar{internvl2} & 2.022 & 0.053 & 52.62 \\
InternVL2 (8B) \citeyearpar{internvl2} & 1.959 & 0.052 & 45.90 \\
VILA-1.5 (8B) \citeyearpar{2024vila} & 2.079 & 0.074 & 44.48  \\
VideoLLaVA (7B) \citeyearpar{2023videollava} & 1.024 & 0.022 & 33.14  \\

\midrule
\multicolumn{4}{c}{\textit{Audio MLLM}} \\ 
\midrule

SALMONN (13B) \citeyearpar{2024salmonn} & 1.758 & 0.053 & 36.48 \\

\bottomrule

\end{tabular}
\end{adjustbox}
\caption{Open-ended QA test on {\ManualBenchmark}. "GPT rank" denotes the average ranking of the model's answers ranked by GPT-4o, "Rouge-L" denotes the average Rouge-L between model's answer and reference answer, and MCQ Acc stands for the model's accuracy(\%) on MCQs.}
\label{tab:Openended Ablation}
\end{table}

\subsection{Evaluating Text Shortcut on Benchmarks}
\begin{table}[tp]
\centering
\begin{adjustbox}{max width=\linewidth}

\begin{tabular}{l!{\vrule width \lightrulewidth}c!{\vrule width \lightrulewidth}c!{\vrule width \lightrulewidth}c!{\vrule width \lightrulewidth}c} 
\toprule
\multicolumn{1}{c|}{\multirow{2}{*}{\textbf{Benchmarks}}} & \multicolumn{2}{c|}{\textbf{Text-Only Accuracy (\%) $\downarrow$}} & \multicolumn{2}{c|}{\textbf{Cycled Accuracy (\%) $\downarrow$}} \\
\cmidrule(lr){2-3}\cmidrule(lr){4-5}
& GPT-4o  & Gemini 1.5 Pro  & GPT-4o & Gemini 1.5 Pro  \\
\midrule

VideoVista \citeyearpar{li2024videovista}& 68.00 &62.63 & 41.25 &38.75\\
TemporalBench \citeyearpar{cai2024temporalbench}& 65.42 &59.37 & 73.57 &52.98 \\
NExT-QA \citeyearpar{2021nextqa} & 47.57 &48.95 & 23.22 &26.37  \\
MLVU \citeyearpar{zhou2024mlvu} & 42.96 &40.98 & 19.38 &12.81  \\
MotionBench \citeyearpar{hong2025motionbench} & 41.24 &36.37 & 16.17 &11.86 \\
Video-MME \citeyearpar{2024videomme} & 41.22 &45.22 & 16.03 &13.05 \\
MVBench \citeyearpar{2024mvbench} & 32.24 &39.03 & 10.84 &5.70  \\ 
Perception Test \citeyearpar{2023perceptiontest}& 29.20 &32.67 & 4.68 &6.02\\
\textbf{\OurBenchmark} & \textbf{25.32} &\textbf{29.61} & \textbf{1.89} &\textbf{4.50}  \\

\bottomrule

\end{tabular}
\end{adjustbox}
\caption{Benchmark Text Shortcut Result. "Text-Only Accuracy" denotes accuracy with text input only. "Cycled Accuracy" denotes accuracy requiring consistent correct answers across four option permutations (ABCD, BCDA, CDAB, DABC). Text shortcut is tested with both GPT-4o and Gemini 1.5 Pro to avoid model bias.}
\label{tab:text leakage}
\end{table}

To demonstrate the prevalence of text shortcut and the effectiveness of our mitigation strategy, we evaluated several video understanding benchmarks alongside {\OurBenchmark}. We used GPT-4o as our test model, assessing its performance in two conditions: (1) using the Text-Only Accuracy (only the question text is given) and (2) using the Cycled Accuracy metric (requiring correct answers across all four permutations).

As shown in Table \ref{tab:text leakage}, existing benchmarks suffer from text shortcut, with both GPT-4o and Gemini-1.5-pro achieving surprisingly high accuracy \textbf{without accessing any video}. In particular, VideoVista had a text-only accuracy of \textbf{68.00}\% with GPT-4o, and the audio-visual benchmark, Video-MME had \textbf{41.22}\%, which are all far above the 25\% accuracy of random guessing. This highlights the ubiquity of the problem and the potential for misleading performance evaluations. In contrast, {\OurBenchmark} \textbf{approaching the expected random guessing accuracy} on four-option multiple-choice questions with text-only input on both GPT-4o and Gemini-1.5-pro. Furthermore, {\OurBenchmark} also achieves a cycled accuracy of \textbf{1.89\%} with GPT-4o that is significantly lower than others by large margins, confirming the effectiveness of our filtering mechanism in mitigating text shortcut. As a result, {\OurBenchmark} provides a more reliable and rigorous assessment of the actual capabilities of video understanding.
\section{Conclusion}

In this paper, we present \textbf{\OurBenchmark}, a text-shortcut-free benchmark designed to advance audio-centric video understanding. {\OurBenchmark} provides a diverse suite of tasks focused on audio understanding and audio-visual alignment, supported by a meticulously curated dataset including both human-annotated and Gemini-augmented subsets.
An extensive evaluation of models highlights the importance of audio information in complete video comprehension. We believe {\OurBenchmark} is valuable for the research community, fostering innovation and progress in audio-centric video understanding and paving the way for more robust and comprehensive multimodal learning systems.

\section*{Limitations}

One major limitation is that our conclusion is only applicable to the current set of open-source audio-only and audio-visual LLMs being slightly outdated, which yields generally worse results than visual-only models due to their deficiency in incorporating audio. We are aware of contemporaneous audio-visual LLMs with implementations yet to be released. We will include their results in our benchmark as soon as they are available.

Another limitation stems from our focus on short videos. While this choice allowed us to effectively evaluate fine-grained audio-centric understanding with existing models, it limits the assessment of long-range temporal reasoning and durational video comprehension abilities. Future work will explore extending our benchmark to longer videos as models improve in these areas.

While we endeavor to ensure a wide range of annotators, it is inevitable that certain annotator bias still exists in data annotation, such as preference towards visual information and bias due to the first language. Moreover, same as many other MCQ-based benchmarks, the MCQ parts of {\OurBenchmark} only explored 4 option cases, leaving other numbers of choices unexplored.

\bibliographystyle{acl_natbib}
\bibliography{main}

\appendix

\section{Example Videos and Tasks}
\label{appendix example videos and tasks}
\begin{figure}[t]
    \centering
    \includegraphics[width=0.6\linewidth]{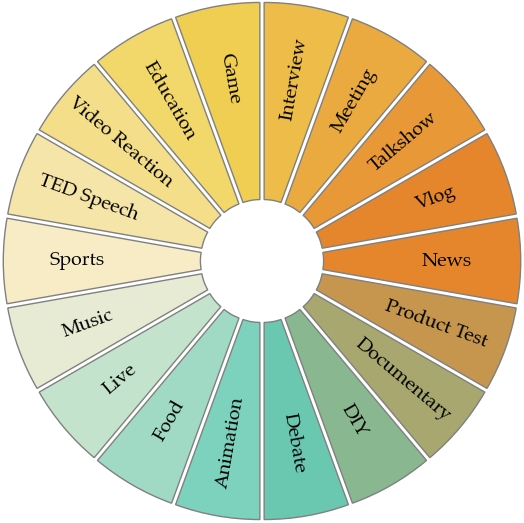}
    \caption{The 18 video domains of {\OurBenchmark} from which the videos are collected.}
    \vspace{-0.3cm}
    \label{fig:video_domain}
\end{figure}

\begin{figure*}[thbp]
    \centering
    \begin{subfigure}{0.32\textwidth}
        \centering
        \includegraphics[width=\linewidth]{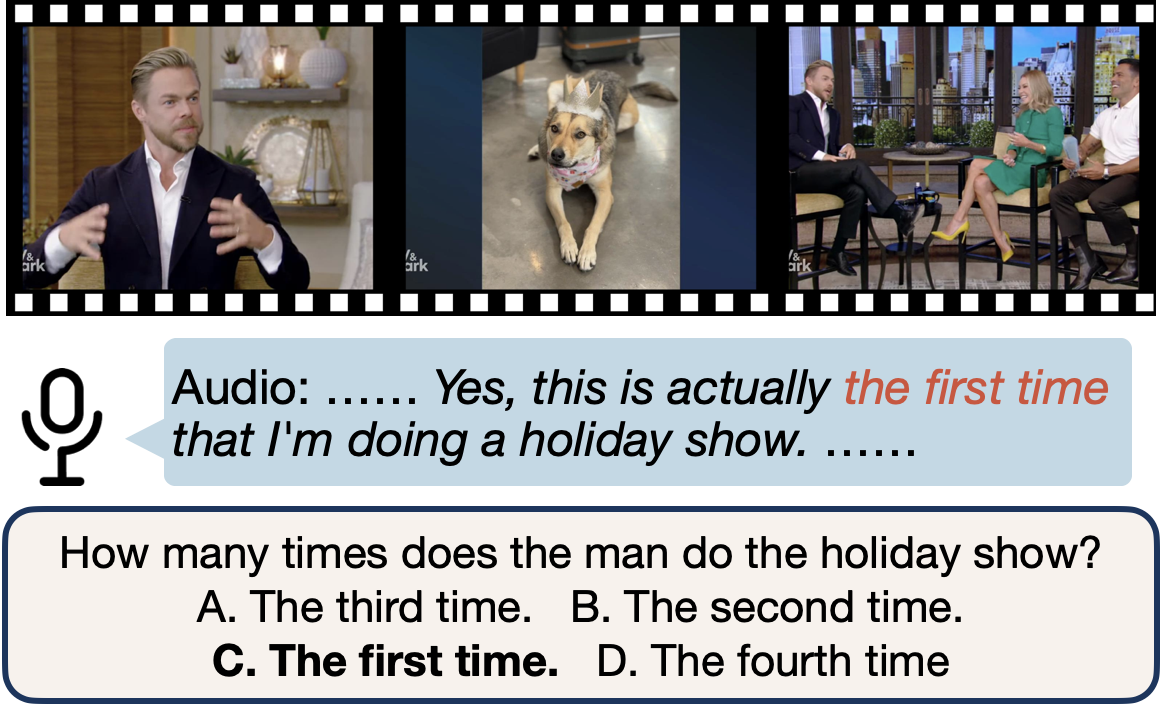}
        \caption{Audio Information Extraction.}
        \label{fig:AIE}
    \end{subfigure}
    \hfill
    \begin{subfigure}{0.32\textwidth}
        \centering
        \includegraphics[width=\linewidth]{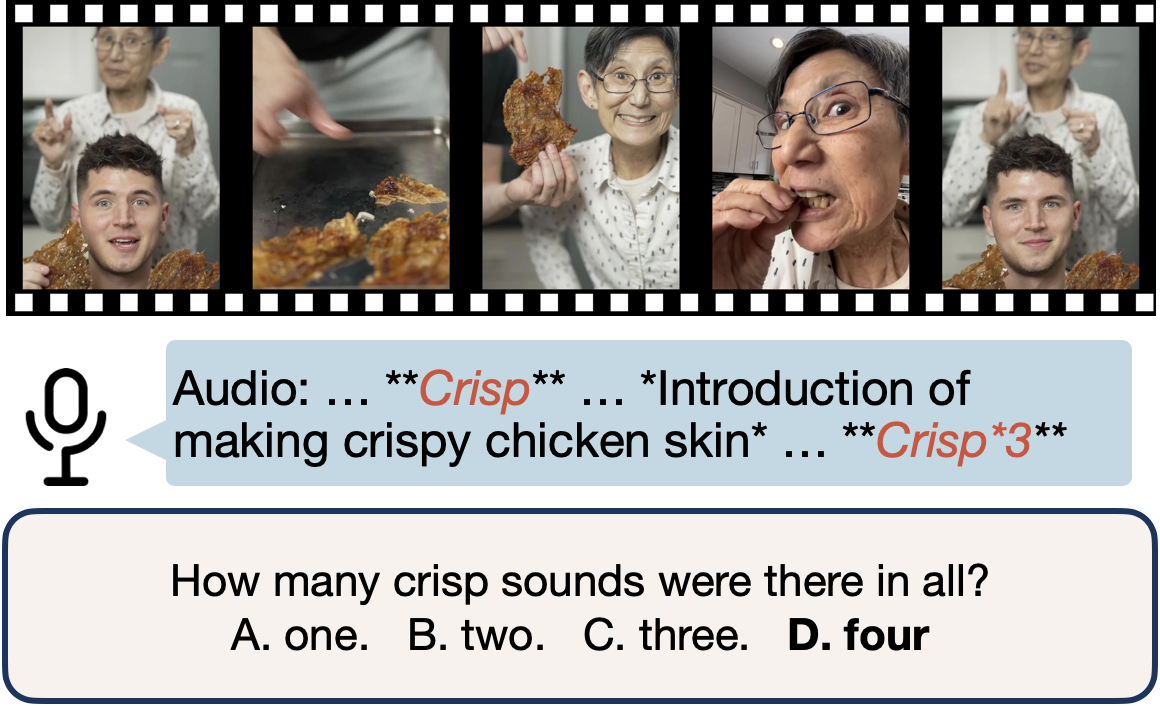}
        \caption{Audio Content Counting}
        \label{fig:ACC}
    \end{subfigure}
    \hfill
    \begin{subfigure}{0.32\textwidth}
        \centering
        \includegraphics[width=\linewidth]{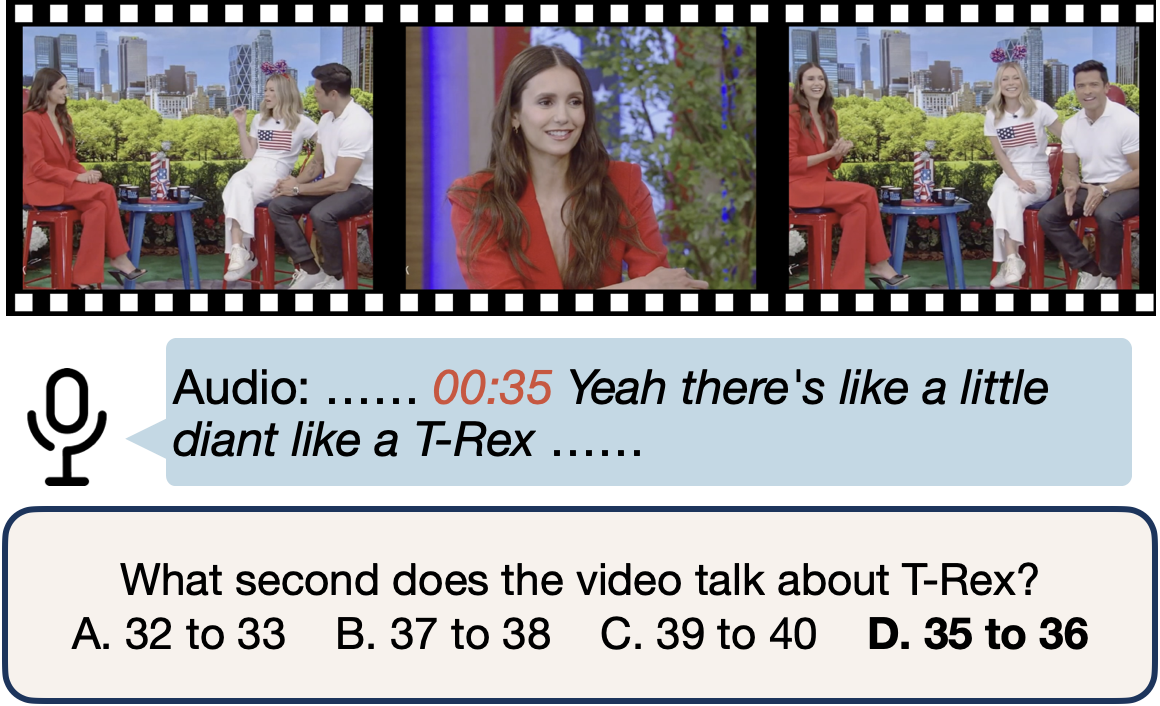}
        \caption{Audio Event Localization}
        \label{fig:AEL}
    \end{subfigure}

    \vskip\baselineskip 

    \begin{subfigure}{0.32\textwidth}
        \centering
        \includegraphics[width=\linewidth]{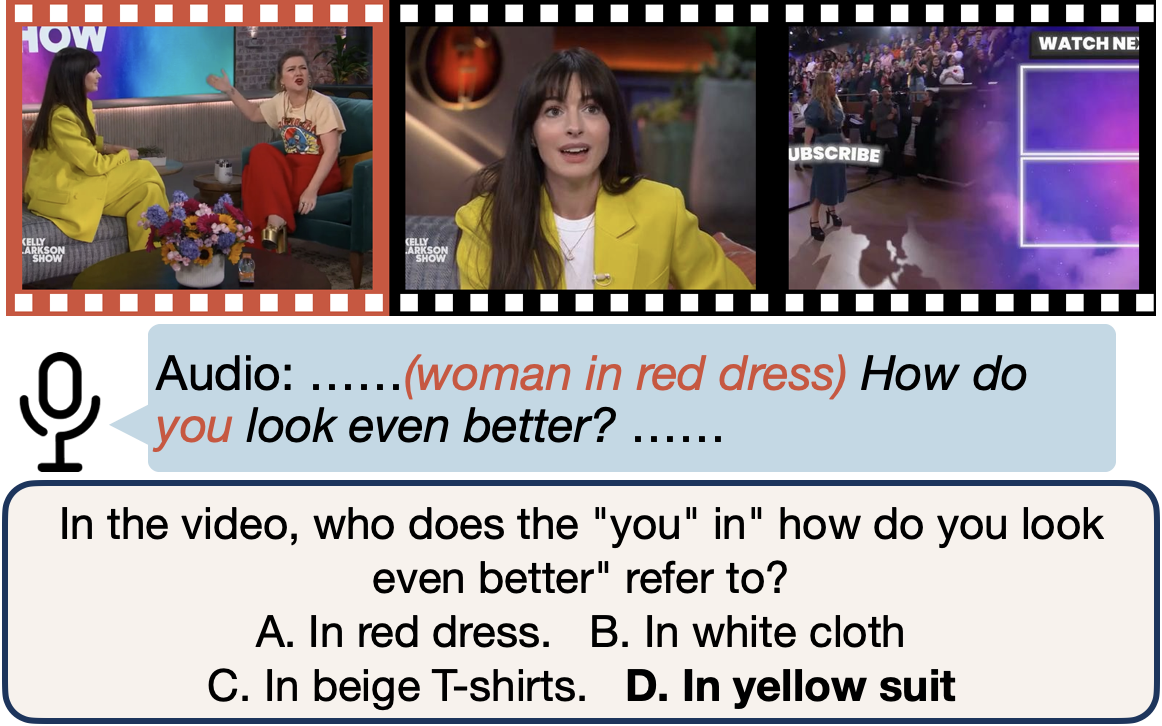}
        \caption{Audio-Visual Character Matching}
        \label{fig:AVCM}
    \end{subfigure}
    \hfill
    \begin{subfigure}{0.32\textwidth}
        \centering
        \includegraphics[width=\linewidth]{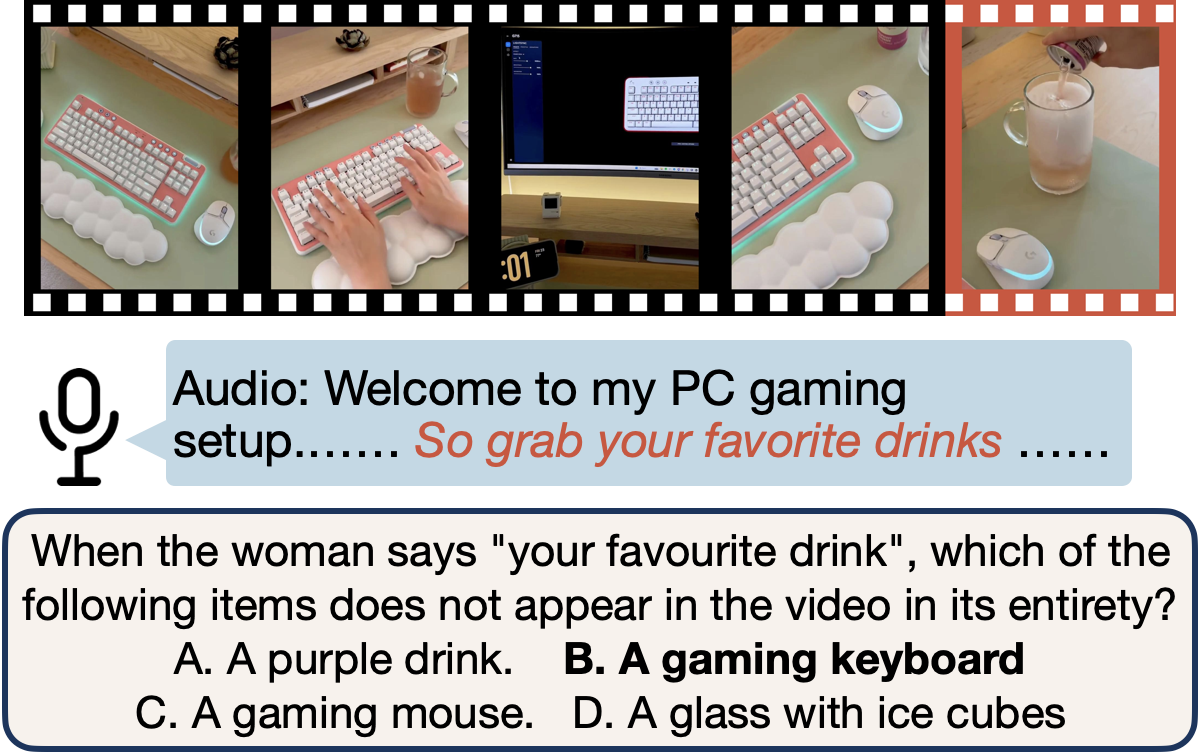}
        \caption{Audio-Visual Object Matching}
        \label{fig:AVOM}
    \end{subfigure}
    \hfill
    \begin{subfigure}{0.32\textwidth}
        \centering
        \includegraphics[width=\linewidth]{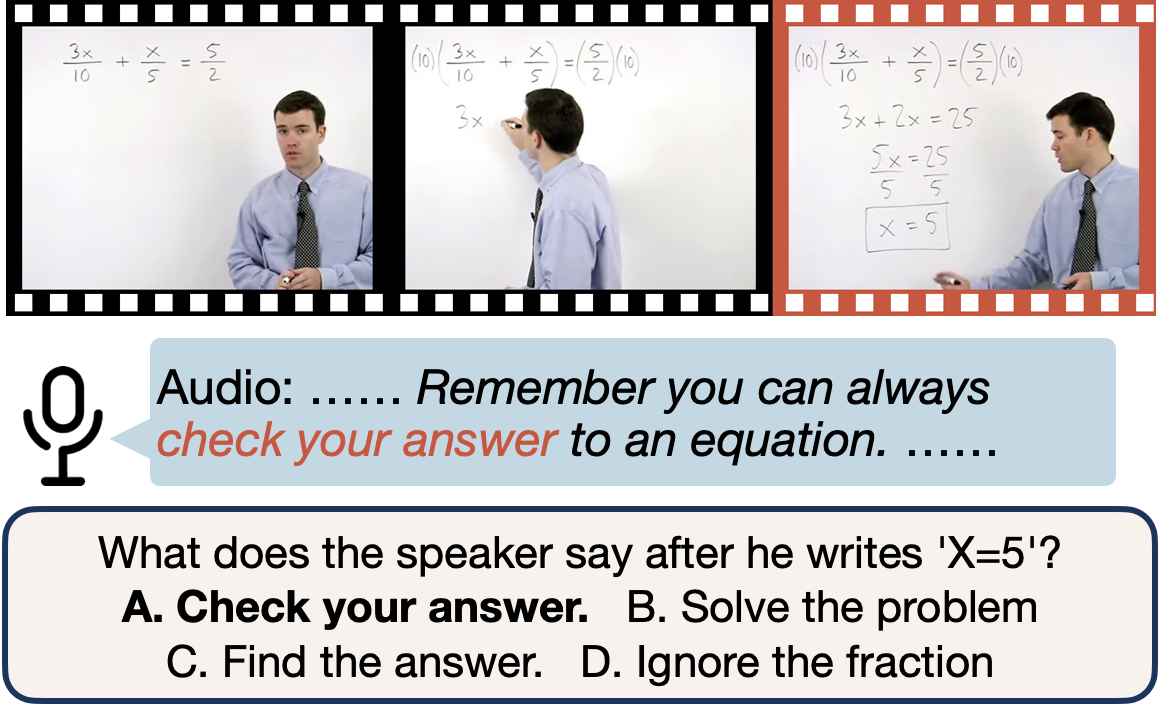}
        \caption{Audio-Visual Text Matching}
        \label{fig:AVTM}
    \end{subfigure}

    \caption{Six MCQ format task examples in \OurBenchmark.}
    \label{fig:six_task_example}
\end{figure*}

This section provides illustrative examples of the six multiple-choice question tasks in {\OurBenchmark}, the video data and corresponding task is in Figure \ref{fig:six_task_example} . Each figure showcases a video frame and the corresponding annotated question with answer options. And 18 audio-centric video domains are shown in Figure \ref{fig:video_domain}.

\section{Case Study on Audio Content Counting and Audio Event Localization Tasks}
\label{appendix:case study}
This section provides an illustrative and detailed case study for the task mentioned difficult for models in the result display section. Two cases on the Audio Content Counting task and one case on the Audio Event Localization task show how visual information is not helpful or even misleading in these tasks, making them harder for models. Answers from outperforming visual-only models and audio-visual models are displayed and analysed for possible difficulties.\newline
\begin{figure*}[thbp]
    \centering
    \begin{subfigure}{0.32\textwidth}
        \centering
        \includegraphics[width=\linewidth, valign=t]{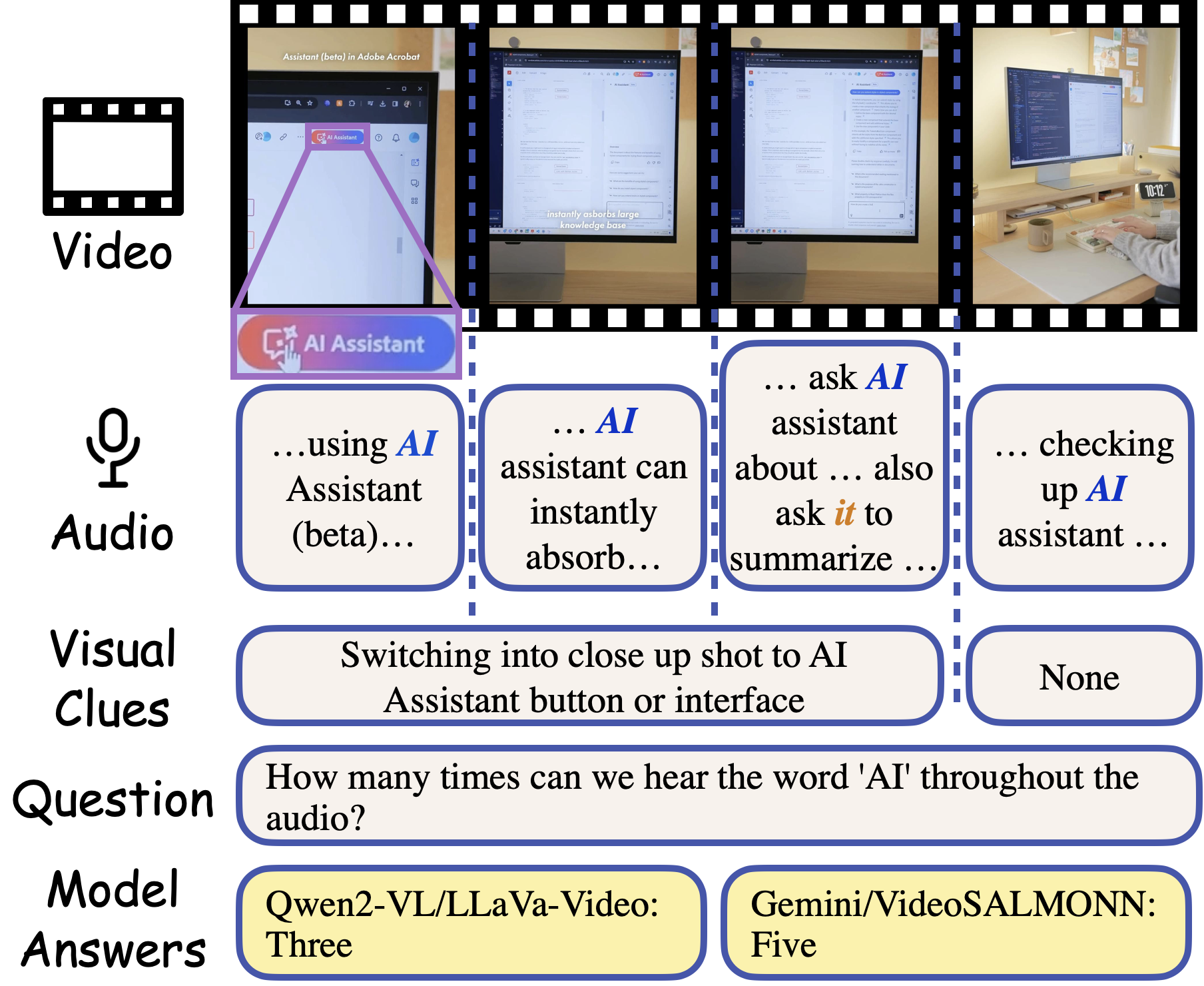}
        \caption{When visual elements misleads models in Audio Content Counting.}
        \label{fig:ACC_case1}
    \end{subfigure}
    \hfill
    \begin{subfigure}{0.32\textwidth}
        \centering
        \includegraphics[width=\linewidth, valign=t]{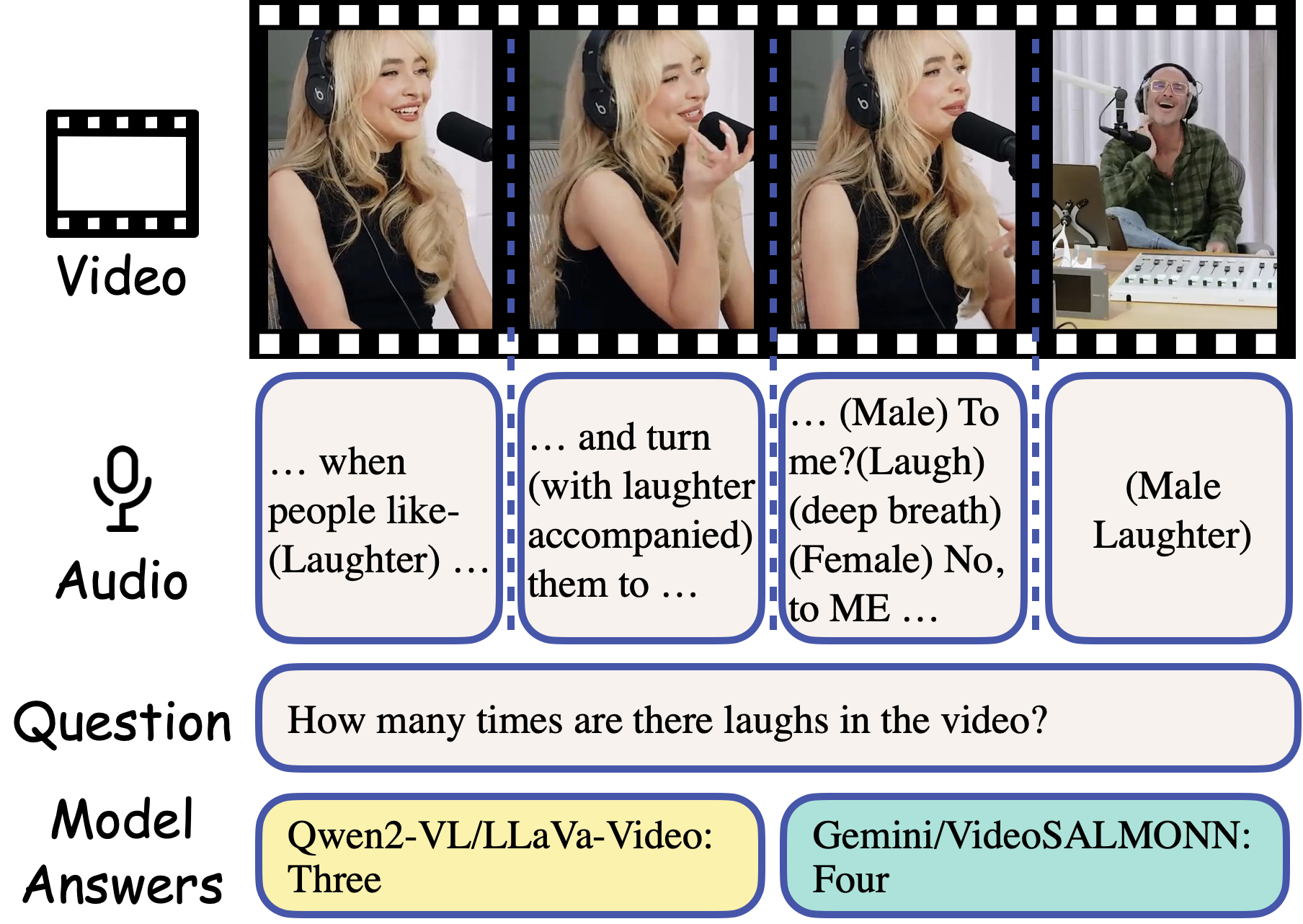}
        \caption{When visual information doesn't help in Audio Content Counting.}
        \label{fig:ACC_case2}
    \end{subfigure}
    \hfill
    \begin{subfigure}{0.32\textwidth}
        \centering
        \includegraphics[width=\linewidth, valign=t]{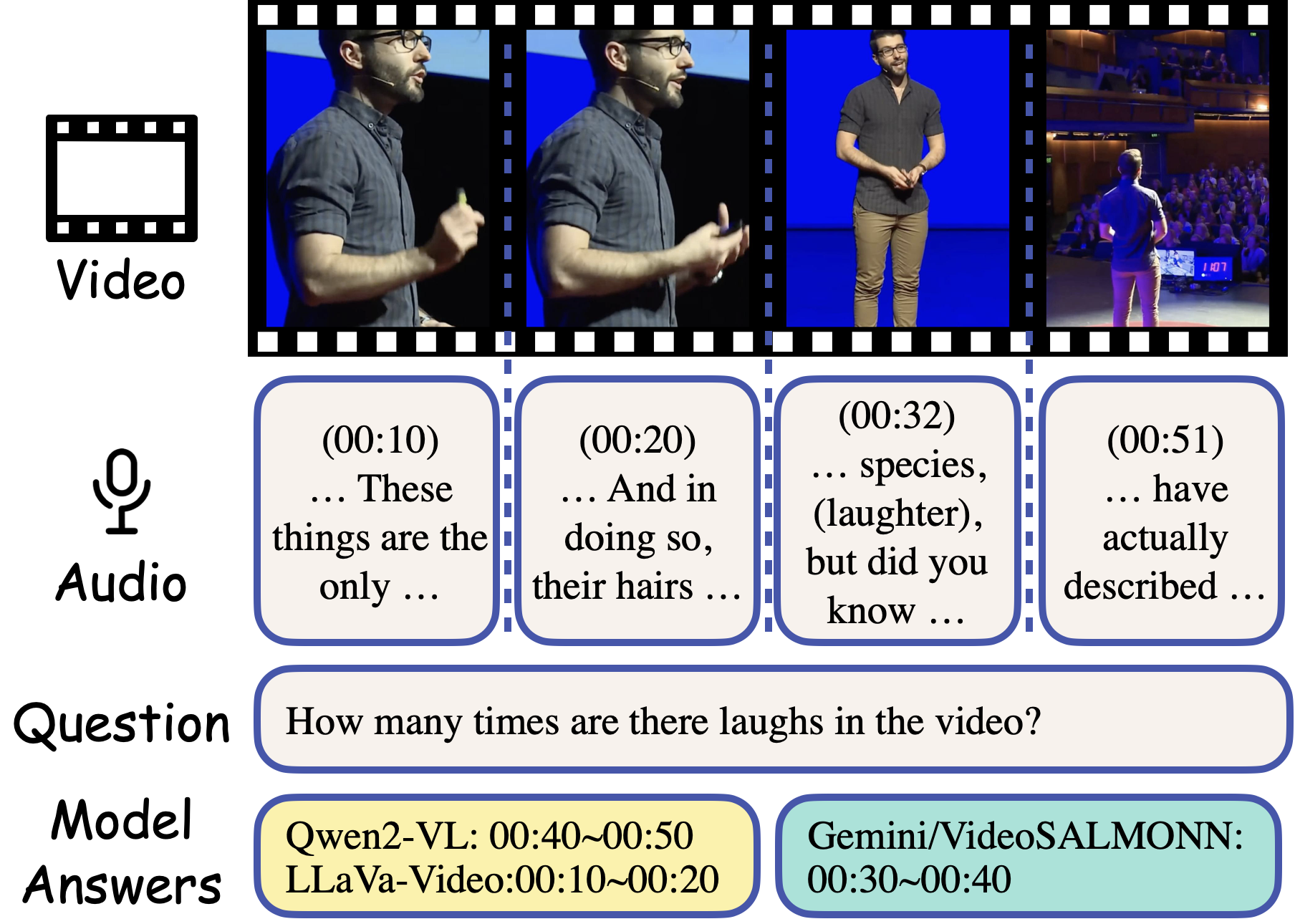}
        \caption{When visual information doesn't help in Audio Event Localization.}
        \label{fig:AEL_case1}
    \end{subfigure}

    \caption{Three cases on models' under-performed tasks.}
    \label{fig:task casestudy}
\end{figure*}
For the case shown in Figure \ref{fig:ACC_case1}, the visual elements provide three relatively separate cues pointing to the word ``AI,'' as shown in the first three example frames in the figure. The last instance includes a fast scene transition, where the visual cue pointing to AI appeared twice in sync with the audio mentions of ``AI'' and ``it.'' Additionally, when ``AI'' is mentioned for the last time in the audio, there is no corresponding visual cue, as shown in the final example frame.\newline
As a result, it can be observed that the best-performing visual models provided answers ``three'' corresponding to the visual cues. In contrast, the audio-visual models produced the answer ``five'', possibly because the strong visual cues and semantic associations in the earlier layers linked the meaning of ``AI'' to the representation of ``it.''\newline
For the case shown in Figure \ref{fig:ACC_case2}, the camera only captured three laughters among all four laughters heard in the audio, as the third laughter from a male interviewer was not captured by the camera, which focused on the female interviewee at the time, presented at the third displayed frame. As a result, models that captures only the visual clues would report three laughters, missing the ones that don't have the camera focused on the laughing characters. Models that can hear, like Gemini 1.5 Pro or video-SALMONN, would be more likely to answer this question correctly.\newline
From the examples above, it can be seen that although the counted audio content generally aligns with the main theme of the video and sometimes corresponds to the visual content, such correspondence provides little help to the audio content counting task. Due to the nature of the counting task, the partial matching does not provide ``partial assistance'', but instead leads to incorrect answers. This makes the models benefit less from the natural, imperfect audio-visual consistency, thereby placing essential demands on the model's audio understanding capabilities.\newline
For Figure \ref{fig:AEL_case1}, the camera does not focus on the subject of ``Laugh'' (which is the audience) at the corresponding moment, as shown in the third frame. As a result, the visual model can only rely on guessing strategies, such as attempting to align with the closest appearance of the audience (like Qwen2-VL), capturing hallucinations, or making random guesses. Only models that really heard the relevant audio are able to locate the event correctly.\newline
The example above illustrates that in the audio event localization task, the audio events involved may be relatively insignificant to the video's theme. As a result, video producers often do not ensure their alignment with visual elements, creating more cases where visual clues alone is insufficient. This further raises the demands on the model’s audio understanding capabilities.

\section{Annotation Information}
Our labelers are provided by a professional outsourcing labeling company, and all of our labelers have bachelor's degree or above and certificates of related professions. For the payment of the labeled person, the labeled person's income is higher than the local minimum income.
\label{appendix:annotation}
\subsection{Annotator Skill Acquisition}
Annotator selection and training were carefully managed to ensure high-quality annotations. Initial screening considered task-specific requirements, matching annotators' skills and experience to appropriate tasks. For example, annotators with sports knowledge were assigned to sports commentary videos, and those with strong listening skills were assigned to audio transcription tasks. This initial optimization aimed to maximize data quality by aligning expertise with task requirements. Following selection, dedicated managers conducted online training sessions, incorporating insights from pilot annotations to address potential challenges and difficulties. Training included detailed explanations of task guidelines and illustrative examples to facilitate annotator comprehension. During the formal annotation phase, managers continued to collaborate with annotators, providing feedback and corrections on initial annotations to further refine their understanding and skills. Additionally, annotators were encouraged to obtain relevant certifications to further enhance their expertise. This multi-faceted approach ensured annotators possessed the necessary skills and knowledge for producing high-quality annotations.

\subsection{Task Allocation}
Task allocation was carefully managed to ensure efficient and balanced workload distribution. The total workload was initially assessed and divided into smaller batches based on the project deadline and the number of available annotators. Tasks were then assigned based on data complexity and annotator skill level. For demanding tasks, such as turn-taking analysis requiring high listening comprehension and repeated video review, batch sizes were adjusted to maintain fairness and prevent annotator overload. The allocation process also included dynamic adjustments to incentivize participation and address workload fluctuations. Incentives, such as increased per-item rates and bonuses for additional batches or overtime work, were implemented to motivate annotators and ensure timely completion. A ``task-grabbing'' mechanism was employed, allowing annotators to claim new batches upon successful completion of previous ones. This prevented individual annotators from taking on excessive workloads and ensured consistent progress. Annotator feedback was actively solicited and incorporated into the allocation strategy. For example, batches identified as unusually difficult were further subdivided and distributed among more annotators. This flexible and responsive approach ensured timely project delivery and consistent annotation quality.

\subsection{Quality Control Procedures}
A rigorous, multi-stage quality control process, involving annotators, quality control specialists, and project managers, was implemented to ensure high-quality annotations. Annotators performed an initial annotation, actively communicating challenges and ambiguities to project managers for clarification and guidance. Unresolved issues were escalated to the researchers for further clarification. Upon completion of the initial annotation phase, dedicated quality control specialists conducted a thorough review, employing methods such as full review, sampling, or cross-validation, as appropriate for the specific task. Annotators then revised their work based on the specialists' feedback. A second round of quality control followed, with any remaining discrepancies adjudicated by a third-party reviewer to ensure final accuracy and consistency. Throughout this process, quality control adhered strictly to the provided guidelines, with specific criteria tailored to each task type (e.g., video retrieval, speech recognition, text annotation). Quality control specialists were selected based on their expertise and skills relevant to the specific task, further ensuring high-quality and consistent results. This meticulous approach guaranteed the accuracy, reliability, and overall quality of the final annotations.

\section{Video Collection Process}
\label{appendix: video collection}
Annotators were tasked with collecting 2700 YouTube videos for each of 18 specified video domains. The following criteria were applied during video collection:

\begin{itemize}[leftmargin=*]
\setlength\itemsep{-0.2em}
    \item \textbf{Audio Content.} Videos need to contain rich audio elements, such as diverse sounds (human speech, environmental sounds, artificial sounds) or substantial spoken dialogue.
    \item \textbf{Language.} Videos were required to be in English.
    \item \textbf{Duration.} Videos were limited to a maximum of two minutes. For longer videos, annotators provided start and end points for a relevant segment within the two-minute limit.
    \item \textbf{Publication Date.} Videos were required to be published in 2023 or later, with preference given to more recent videos (ideally from 2024) to minimize overlap with model training data.
    \item \textbf{Subtitles.} Videos could not contain manually added subtitles or captions.
    \item \textbf{Video Domains.} The 18 target video domains were: Game, Interview, Meeting, Talkshow, Vlog, News, Product Test, Documentary, DIY, Debate, Animation, Food, Live, Music, Sports, TED Speech, Video Reaction, and Education. Specific definitions and examples were provided for each video domain to ensure consistency.
\end{itemize}

\section{Annotation Guidelines and Examples}
\label{appendix: guidelines and examples}
This section details the annotation process for creating question-answer pairs for the {\ManualBenchmark}. Annotators were provided with 700 videos and tasked with creating three question-answer pairs per video, totaling 1,734 annotations. The annotation process followed these steps:

(1) Video Comprehension: Annotators thoroughly watched each video to understand its content.

(2) Task Selection: Annotators reviewed the six multiple-choice format task types (listed below) and selected the three most relevant tasks for each video. Relevance was determined by whether the video's content could be used to create a meaningful question within the chosen task type.

(3) Question-Answer Design: Annotators designed a multiple-choice question and four answer options (A, B, C, and D) for each selected task type, based on the video content and the specific task guidelines. Only one answer option could be correct.

(4) Rationality Check: Annotators performed a rationality check on each question-answer pair to ensure it met the following criteria:
\begin{itemize}[leftmargin=*]
\setlength\itemsep{-0.2em}
    \item \textbf{Audio Dependence.} The question could not be answered correctly by watching the video without audio.
    \item \textbf{Clear Distinctions.} The correct answer needed to be clearly distinguishable from the incorrect options, avoiding ambiguity.
    \item \textbf{No External Knowledge.} Questions and options could not rely on common sense knowledge or information beyond the video's content.
    \item \textbf{Data Formatting.} Annotations were recorded in an Excel spreadsheet with the following columns: video id, URL, task type, question, option A, option B, option C, option D, and answer.
\end{itemize}

To aid annotators in understanding and applying each task type, we provided two annotated examples with accompanying videos for each task. These examples served as templates and illustrated best practices for question-answer design. The task definitions provided for annotators are as follows:
\begin{itemize}[leftmargin=*]
\setlength\itemsep{-0.2em}
    \item \textbf{Audio Information Extraction.} This task focuses on extracting specific details conveyed through spoken language in the video. Questions should be direct, clear, and objective, targeting factual information from the audio. Answer options should be relevant, concise, and clearly distinguishable, with distractor options designed to assess the depth of audio comprehension. 
    \item \textbf{Audio Content Counting.} This task requires counting the occurrences of specific audio events within the video. Questions should clearly define the target event and the scope of counting (e.g., the entire video or a specific segment). Answer options should provide a reasonable range of numerical choices with appropriate intervals. Distractor options should be numerically close to the correct answer to test counting precision.
    \item \textbf{Audio Event Localization.} This task involves pinpointing the precise time a specific audio event occurs. Questions should clearly identify the target event and request its temporal location within the video. Answer options should offer distinct time ranges with reasonable granularity (seconds precision), using the format ``time1 to time2''. Distractor options should test the model's ability to precisely locate the event. 
    \item \textbf{Audio-Visual Character Matching.} This task requires matching spoken information from a character to their visual appearance. Questions should link a character's utterance to their visual characteristics. Answer options should describe different characters or visual features present in the video. Distractor options should include other characters or similar visual features to test the model's matching accuracy.
    \item \textbf{Audio-Visual Object Matching.} This task focuses on matching spoken information to objects or scenes within the video. Questions should link spoken descriptions or references to the corresponding visual elements. Answer options should describe different objects or scenes present in the video. Distractor options should include other objects or similar scenes to evaluate the model's ability to connect audio and visual elements.
    \item \textbf{Audio-Visual Text Matching.} This task involves matching spoken information to on-screen text (OCR). Questions should connect spoken words or phrases to specific text appearing in the video, often specifying the timing and location of the text. Answer options should offer a variety of OCR text strings or combinations. Distractor options should include similar-looking text or text from different locations to test the model's ability to precisely match audio and text. 
\end{itemize}

\section{Gemini-Generated Annotation Review Process}
\label{appendix: Gemini refinement}
This section details the human review process for the 9,875 Gemini-generated question-answer pairs. Annotators were tasked with ensuring the generated content met specific quality criteria, focusing on question relevance, option validity, answer accuracy, and adherence to video content without reliance on external knowledge or common sense reasoning. The review process for each question-answer pair involved the following steps:

(1) Video Review: Annotators watched the entire video or the specified segment (for longer videos with provided start and end points) to fully understand the content.

(2) Task Type Verification: Annotators verified that the assigned task type aligned with the video content. The definitions of the six task types (Audio Character Matching, Audio Object Matching, Audio OCR Matching, Audio Event Location, Audio Content Counting, and Audio Information Extraction) were provided to the annotators (see Section \ref{appendix: guidelines and examples} for definitions).

(3) Question Evaluation: Annotators assessed the question's relevance to both the task type and the video content, ensuring it could not be answered solely through text reading or common sense reasoning.

(4) Option Evaluation: Annotators checked the validity of all four answer options.

Our human review stage was critical for mitigating biases and ensuring annotation quality. Approximately 40\% of the 9,875 generated QA pairs required modification, ranging from minor rephrasing to complete rewriting. Common categories of issues encountered during the review process are detailed below, along with illustrative examples:

\begin{itemize}
    \item \textbf{Task Type Mismatch (20\%):} Gemini generated questions that did not align with the intended task type. For example:
        \begin{itemize}
            \item Intended Task Type: Audio Visual Character Matching (requiring matching characters based on both audio and visual information)
            \item Gemini-Generated Question: \texttt{``What is the narrator wearing?'', Options: A) Red shirt, B) Black suit, C) Red suit, D) Yellow suit}
            \item Gemini-Generated Answer: \texttt{D}
            \item Explanation of Error: The ``Audio Visual Character Matching'' task requires identifying a character based on both their audio and visual presence. However, in this video, the narrator is only heard and never seen on screen. Thus, the generated question about the narrator's visual appearance is inherently flawed and incompatible with both the task's requirements and the video content itself.
        \end{itemize}

    \item \textbf{Irrelevant to Video Content (15\%):} Questions were generated that did not pertain to the content of the video. For example:
        \begin{itemize}
            \item Intended Task Type: Audio Visual Object Matching
            \item Gemini-Generated Question: \texttt{``What is the man talking about when he says, `This car is 10 years old and you would not know that by looking at it'?'', Options: A) A dryer, B) Clothes, C) A car, D) Plastic wrap}
            \item Gemini-Generated Answer: \texttt{C}
            \item Explanation of Error: The generated question can be answered solely by interpreting the provided quote. The video content is entirely superfluous.
        \end{itemize}

    \item \textbf{Multiple Correct Answers (15\%):} Questions were formulated in a way that allowed multiple valid answers. For example:
        \begin{itemize}
            \item Intended Task Type: Audio Event Localization
            \item Gemini-Generated Question: \texttt{``When does the speaker say the word `pumpkin'?'', Options: A) 0:00-0:05, B) 0:05-0:10, C) 0:10-0:15, D) 0:15-0:20}
            \item Gemini-Generated Answer: \texttt{D}
            \item Explanation of Error: The speaker says "pumpkin" in both time ranges C (0:10-0:15) and D (0:15-0:20). Thus, the question as initially formulated has two correct answers.
        \end{itemize}

    \item \textbf{Incorrect Correct Answer (40\%):} The answer provided by Gemini as "correct" was factually incorrect. For example:
        \begin{itemize}
            \item Intended Task Type: Audio Content Counting
            \item Gemini-Generated Question: \texttt{``How many times does the phone vibrate in the video?'', Options: A) 1, B) 2, C) 3, D) 4}
            \item Gemini-Generated Answer: \texttt{A}
            \item Explanation of Error: Gemini's provided answer (``A: 1'') is factually incorrect. The phone does not vibrate in the video.
        \end{itemize}

    \item \textbf{Other Issues (10\%):} This category encompasses issues such as overly simplistic questions, lack of distractor quality, generated gibberish, etc.
\end{itemize}

\section{Data Availability and Copyright}
Our benchmark {\OurBenchmark} is constructed using publicly available videos from YouTube. We adhere to YouTube's copyright policies and terms of service. {\OurBenchmark} contains only links to these public videos; we do not host or distribute copies of the video content itself. This ensures compliance with copyright regulations and allows for easy access to the data while respecting intellectual property rights. Users of {\OurBenchmark} are responsible for complying with YouTube's terms of service when accessing the linked videos.

\section{Use of Gemini 1.5 Pro and GPT-4o}
This section clarifies the use of Gemini 1.5 Pro and GPT-4o within our research and our adherence to their respective terms of service.\footnote{Gemini Terms of Service: \url{https://ai.google.dev/gemini-api/terms\#use-generated}. GPT-4o Terms of Service: \url{https://openai.com/policies/business-terms/}.}Our use of these models is strictly for non-commercial, research purposes and does not involve any commercial applications or profit generation.

Gemini 1.5 Pro: Gemini 1.5 Pro was employed solely for data augmentation purposes, specifically to generate a larger set of question-answer pairs for our benchmark. In accordance with Google's terms of service, we acknowledge that Google retains all rights to generate similar content for other users. We do not claim ownership over the Gemini-generated content. We also acknowledge our responsibility for the use of this generated content, both within our research and by anyone with whom we share the benchmark data. While we have used discretion in relying on the generated content, users of {\OurBenchmark} should be aware of its automated origin. As required by the API Terms, we will comply with applicable laws regarding the use of generated content, including providing attribution when necessary.

GPT-4o: GPT-4o was utilized exclusively for evaluation purposes, specifically to assess the performance of various models on our benchmark. Our use of GPT-4o for evaluation is in compliance with its terms of service.

\section{Model Prompting Details}
This section provides the specific prompt template used in the experiments. \newline
Prompt \ref{alg:1} is used for evaluating models on the multiple-choice question tasks in {\OurBenchmark}. This prompt was provided to each model.\newline
Prompt \ref{alg:2} is used for evaluating models on the open-ended question tasks in {\ManualBenchmark}. This prompt was provided to each model.\newline
Prompt \ref{alg:3} is used for prompting GPT-4o to provide a ranking to the answer that models generated for questions in {\ManualBenchmark}. This prompt was provided to GPT-4o to create a ranking for each answer.\newline
\begin{algorithm*}[t]
\caption{Unified prompt for testing.}
\label{alg:1}
\begin{lstlisting}[style=algorithmstyle]
Select the best answer to the following multiple-choice question based on the video.
(*@\textbf{\textit{data["question"]}}@*)

A. (*@\textbf{\textit{data["option\_A"]}}@*)
B. (*@\textbf{\textit{data["option\_B"]}}@*)
C. (*@\textbf{\textit{data["option\_C"]}}@*)
D. (*@\textbf{\textit{data["option\_D"]}}@*)

Respond with only the letter (A, B, C, or D) of the correct option. The best answer is:
\end{lstlisting}
\end{algorithm*}
\begin{algorithm*}[t]
\caption{Unified prompt for open-ended ablation.}
\label{alg:2}
\begin{lstlisting}[style=algorithmstyle]
Answer the question based on the video: (*@\textbf{\textit{data["question"]}}@*)
\end{lstlisting}
\end{algorithm*}
\begin{algorithm*}[t]
\caption{Unified prompt for GPT-4o to rank the open-ended answer.}
\label{alg:3}
\begin{lstlisting}[style=algorithmstyle, breaklines=true]
You are a helpful assistant. Given the question, predicted answer and the reference answer, rank the predicted answer from 1 to 5 by how similar it is compared with the reference answer. 1 means the predicted answer has nothing to do with the reference answer or conflicts with the reference answer, and 5 means the predicted answer and the reference answer expresses the same thing. Provide only the number.
    The question is: (*@\textbf{\textit{question}}@*)
    The predicted answer is: (*@\textbf{\textit{model\_answer}}@*)
    The reference answer is: (*@\textbf{\textit{reference\_answer}}@*)
    Your ranking is:
\end{lstlisting}
\end{algorithm*}
\section{Annotator Biases}
\label{appendix: annotation bias}
While we implemented rigorous quality control measures, it's important to acknowledge potential biases in human-generated annotations. Analysis of the {\ManualBenchmark} revealed instances where annotations exhibited a stronger reliance on visual information than intended. This section presents a few examples to illustrate this bias and highlight the challenges of achieving truly audio-centric annotation. These examples underscore the importance of careful annotation guidelines, thorough training, and robust quality control processes to mitigate such biases. Furthermore, they motivate the use of semi-automatic annotation methods, such as our Gemini-based approach, to further diversify the data and reduce the impact of human biases.

\begin{lstlisting}[style=jsonstyle]
{
    "video_id": 1,
    "url": "https://www.youtube.com/shorts",
    "video_type": "vlog",
    "task_type": "Audio Event Localization",
    "question": "During which time period, the video does not record the front face of the woman in the plaid shirt?",
    "option_A": "0s-15s.",
    "option_B": "16s-30s.",
    "option_C": "31s-45s.",
    "option_D": "46s-60s.",
    "answer": "C"
},
{
    "video_id": 2,
    "url": "https://www.youtube.com/watch",
    "video_type": "animation",
    "task_type": "Audio-Visual Object Matching",
    "question": "What is on the boy's cloth?",
    "option_A": "A boy.",
    "option_B": "An animal.",
    "option_C": "A girl.",
    "option_D": "A creature like fox.",
    "answer": "D"
},
{
    "video_id": 3,
    "url": "https://www.youtube.com/watch",
    "video_type": "education",
    "task_type": "Audio-Visual Object Matching",
    "question": "What scene does the video show at the beginning?",
    "option_A": "Home.",
    "option_B": "Shop.",
    "option_C": "Supermarket.",
    "option_D": "Theatre.",
    "answer": "C"
},
{
    "video_id": 4,
    "url": "https://www.youtube.com/shorts",
    "video_type": "music",
    "task_type": "Audio-Visual Object Matching",
    "question": "What color bass does the guy who plays the bass at the end of the video have?",
    "option_A": "Yellow and pink.",
    "option_B": "Black and white.",
    "option_C": "Grey and pink.",
    "option_D": "Pink and green.",
    "answer": "B"
}
\end{lstlisting}

\section{Transcription Generation with Gemini}
\label{appendix: Gemini transcription}
This section provides examples of the detailed transcriptions generated using Gemini 1.5 Pro for {\OurBenchmark}. These transcriptions include speaker identification, corresponding ASR transcripts, and start/end timestamps for each utterance. The inclusion of detailed transcriptions allows us to investigate the extent to which textual representations of audio can substitute for the original audio in video understanding tasks.

\begin{lstlisting}[language=json, caption=Example Transcription, captionpos=b, basicstyle=\ttfamily\footnotesize, breaklines=true, frame=single]
[
    {
        "start_time": 0.0,
        "end_time": 2.0,
        "speaker": "Speaker 1",
        "characteristic": "female with long brown hair, wearing a blue dress with flowers",
        "transcript": "I heard that you brought your parents to the Oscars was that fun?"
    },
    {
        "start_time": 2.0,
        "end_time": 6.0,
        "speaker": "Speaker 2",
        "characteristic": "female with blonde hair, wearing a blue jacket with flowers",
        "transcript": "I did I did my dad actually called me and sort of invited himself."
    }
]
\end{lstlisting}
\end{document}